\pdfoutput=1
\def\BottleneckPreprint{1}
\def\BottleneckPdfAuthor{Zichuan Xu, Yuyang Dong, Makoto Onizuka, Chuan Xiao}
\def\BottleneckAuthorBlock{%
Zichuan Xu$^{1}$ \quad Yuyang Dong$^{2}$ \quad
Makoto Onizuka$^{1}$ \quad Chuan Xiao$^{1,3}$\\[0.25em]
{\normalfont\small $^{1}$University of Osaka \quad $^{2}$Initial S \quad $^{3}$Nagoya University}\\
{\normalfont\small\texttt{\{xu.zichuan,onizuka,chuanx\}@ist.osaka-u.ac.jp}}\\
{\normalfont\small\texttt{dongyuyang@initial-s.com}}}

\documentclass[11pt]{article}

\let\BBrealbibliographystyle\bibliographystyle
\renewcommand{\bibliographystyle}[1]{}
\ifdefined\BottleneckPreprint
  \usepackage[preprint]{acl}
\else
  \usepackage[review]{acl}
\fi
\let\bibliographystyle\BBrealbibliographystyle
\usepackage{times}
\usepackage{latexsym}
\usepackage{amsmath,amssymb}
\usepackage{array,booktabs,graphicx,microtype}
\usepackage{pifont}
\usepackage{placeins}
\usepackage[T1]{fontenc}
\usepackage[utf8]{inputenc}
\usepackage{paralist}

\makeatletter\@secpenalty=0\makeatother

\providecommand{\Description}[1]{}

\newcommand{\figref}[1]{\mbox{Figure~\ref{#1}}}

\newcommand{\HolmEq}[1]{$=#1$}

\newcommand{\CoreModels}{21}
\newcommand{\AuxModels}{20}
\newcommand{\StructuralModels}{17}
\newcommand{\RegisteredStructuralModels}{2}
\newcommand{\AdditionalStructuralModels}{15}
\newcommand{\ConcurrentMenuModels}{12}
\newcommand{\ExternalRankModels}{5}
\newcommand{\HistoricalMenuModels}{5}
\newcommand{\Organizations}{7}

\newcommand{\TotalRows}{26{,}576}

\newcommand{\SentinelRows}{1{,}440}

\newcommand{\StructuralRows}{3{,}672}

\newcommand{\ConcurrentMenuRows}{1{,}152}
\newcommand{\ConcurrentMenuBlocks}{384}

\newcommand{\ExactVTSRows}{20{,}408}

\newcommand{\SemanticFamilies}{36}
\newcommand{\Intents}{12}

\newcommand{\HoldoutIntents}{24}
\newcommand{\DistractorExecutions}{13{,}158}

\newcommand{\SuccessResponseEpisodes}{9{,}841}
\newcommand{\SuccessResponseTokensMedian}{152}
\newcommand{\SuccessResponseTokensPNN}{476}

\newcommand{\CompletionTokenCap}{1{,}024}
\newcommand{\EpisodeTokenBudget}{128{,}000}

\newcommand{\SchemaBytesVEight}{3{,}481}

\newcommand{\SchemaBytesVOneTwoEight}{55{,}176}
\newcommand{\InfrastructureCapacityConditions}{8}
\newcommand{\InfrastructureDiscoveryConditions}{1}
\newcommand{\InfrastructureConditions}{9}
\newcommand{\CoreVTS}{82.1}

\newcommand{\VisibleEffect}{\ensuremath{-}12.3}

\newcommand{\VisibleEffectCILow}{\ensuremath{-}14.6}
\newcommand{\VisibleEffectCIHigh}{\ensuremath{-}9.9}
\newcommand{\VisibleHolmP}{\ensuremath{<}.001}
\newcommand{\VisibleConfigCILow}{\ensuremath{-}19.8}
\newcommand{\VisibleConfigCIHigh}{\ensuremath{-}5.7}
\newcommand{\VisibleConfigHolmP}{.011}

\newcommand{\VisibleDegradedFamilies}{34}

\newcommand{\WorkflowEffectSigned}{+0.8}
\newcommand{\WorkflowEffectCILow}{\ensuremath{-}1.2}
\newcommand{\WorkflowEffectCIHigh}{2.8}

\newcommand{\WorkflowConfigCILow}{\ensuremath{-}4.6}
\newcommand{\WorkflowConfigCIHigh}{6.5}
\newcommand{\WorkflowConfigHolmP}{1.000}

\newcommand{\CompactEffect}{\ensuremath{-}3.7}

\newcommand{\CompactEffectCILow}{\ensuremath{-}5.7}
\newcommand{\CompactEffectCIHigh}{\ensuremath{-}1.9}
\newcommand{\CompactHolmP}{.003}
\newcommand{\CompactConfigCILow}{\ensuremath{-}5.0}
\newcommand{\CompactConfigCIHigh}{\ensuremath{-}2.5}
\newcommand{\CompactConfigHolmP}{\ensuremath{<}.001}

\newcommand{\CompactExactIntentP}{.0098}
\newcommand{\CompactExactIntentHolmP}{.068}

\newcommand{\InteractionEffectSigned}{+3.0}
\newcommand{\InteractionEffectCILow}{0.2}
\newcommand{\InteractionEffectCIHigh}{5.7}
\newcommand{\InteractionHolmP}{.202}

\newcommand{\ContrastPairs}{756}
\newcommand{\BootstrapDraws}{10{,}000}
\newcommand{\BootstrapPFloor}{.0008}

\newcommand{\InteractionIndependenceBenchmark}{0.1}
\newcommand{\InteractionFloorPatternShare}{68.9}
\newcommand{\InteractionFloorUnits}{31}
\newcommand{\InteractionNonFloorEffect}{2.0}
\newcommand{\InteractionNonFloorCILow}{\ensuremath{-}0.3}
\newcommand{\InteractionNonFloorCIHigh}{4.1}

\newcommand{\NonMonotoneShare}{13.2}
\newcommand{\NonMonotoneUnits}{200}
\newcommand{\InteractionUnits}{1{,}512}

\newcommand{\RoutingGain}{2.4}
\newcommand{\RoutingGainCILow}{1.1}
\newcommand{\RoutingGainCIHigh}{3.8}
\newcommand{\RoutingPairs}{3{,}024}
\newcommand{\RoutingTokenSaving}{4{,}699}
\newcommand{\RoutingSchemaExposureSaving}{47.7}

\newcommand{\OrgBalancedVisibleEffect}{\ensuremath{-}15.3}
\newcommand{\OrgBalancedWorkflowEffect}{3.8}
\newcommand{\OrgBalancedCompactEffect}{\ensuremath{-}4.7}

\newcommand{\TruncationRateVEight}{1.6}
\newcommand{\TruncationRateVOneTwoEight}{13.6}
\newcommand{\TruncatedFailuresVEight}{48}
\newcommand{\TruncatedFailuresVOneTwoEight}{411}

\newcommand{\TruncatedBeforeFirstCallVOneTwoEight}{391}
\newcommand{\VisiblePairs}{3{,}024}
\newcommand{\PairsLost}{472}
\newcommand{\PairsLostTruncated}{343}

\newcommand{\PairsGained}{101}
\newcommand{\PairsLostPoints}{15.6}
\newcommand{\PairsLostTruncatedPoints}{11.3}

\newcommand{\VisibleBoundLower}{\ensuremath{-}13.9}
\newcommand{\VisibleBoundUpper}{1.3}
\newcommand{\CommonResolutionLower}{\ensuremath{-}12.3}
\newcommand{\CommonResolutionUpper}{\ensuremath{-}0.3}
\newcommand{\NonTruncatedPairedEffect}{\ensuremath{-}1.5}
\newcommand{\NonTruncatedPairs}{2{,}595}

\newcommand{\VisibleTruncationPearson}{\ensuremath{-}0.95}
\newcommand{\VisibleTruncationSpearman}{\ensuremath{-}0.82}
\newcommand{\ZeroTruncationConfigs}{8}
\newcommand{\ZeroTruncationEffectMin}{\ensuremath{-}2.1}
\newcommand{\ZeroTruncationEffectMax}{2.8}
\newcommand{\WorkflowArmTruncationGap}{41}
\newcommand{\DeepSeekFlashVisibleEffect}{\ensuremath{-}18.1}
\newcommand{\DeepSeekFlashTruncation}{0.7}

\newcommand{\TotalFailures}{3{,}725}
\newcommand{\FailureRate}{18.3}
\newcommand{\CompletionCapFailures}{1{,}068}
\newcommand{\CompletionCapShare}{28.7}
\newcommand{\BudgetLimitFailures}{1{,}863}
\newcommand{\BudgetLimitShare}{50.0}
\newcommand{\InfeasibilityFailures}{477}
\newcommand{\InfeasibilityShare}{12.8}

\newcommand{\WrongConclusionFailures}{204}
\newcommand{\WrongConclusionShare}{5.5}

\newcommand{\OtherContractFailures}{113}
\newcommand{\OtherContractShare}{3.0}

\newcommand{\HarnessLimitShare}{91.5}
\newcommand{\HarnessLimitShareNoInfeasibility}{78.7}
\newcommand{\FirstResponseTruncatedFailures}{838}

\newcommand{\CapacityCellHarnessLimitShare}{86.9}
\newcommand{\ThirtyCallFailures}{896}

\newcommand{\StageSelection}{849}

\newcommand{\ZeroTokenRows}{1{,}118}
\newcommand{\ZeroTokenAvailabilityRows}{1{,}113}
\newcommand{\ZeroTokenFailureRows}{5}

\newcommand{\DiscoveryRows}{4{,}536}
\newcommand{\DiscoveryUnavailable}{1{,}113}
\newcommand{\DiscoveryUnavailableShare}{24.5}

\newcommand{\DiscoveryAgentFailures}{30}

\newcommand{\AvailabilityFlatLarge}{19.4}
\newcommand{\AvailabilityRouterLarge}{77.8}
\newcommand{\RetrievalConditionalSuperior}{1}
\newcommand{\RetrievalConditionalCases}{7}

\newcommand{\DiscoveryRouterLoss}{\ensuremath{-}21.6}
\newcommand{\DiscoveryRouterLossCILow}{\ensuremath{-}40.1}
\newcommand{\DiscoveryRouterLossCIHigh}{\ensuremath{-}5.0}

\newcommand{\DiscoveryClusters}{18}
\newcommand{\FrontierModels}{20}
\newcommand{\FrontierFamilies}{4}
\newcommand{\FrontierEpisodesPerLevel}{80}
\newcommand{\FrontierVTSAtTwo}{91.2}

\newcommand{\FrontierVTSAtSix}{87.5}

\newcommand{\FrontierVTSAtFourteen}{85.0}

\newcommand{\FrontierVTSAtThirty}{0.0}

\newcommand{\FrontierSixMinusTwo}{\ensuremath{-}3.8}
\newcommand{\FrontierSixMinusTwoCILow}{\ensuremath{-}8.8}
\newcommand{\FrontierSixMinusTwoCIHigh}{2.5}
\newcommand{\FrontierFourteenMinusTwo}{\ensuremath{-}6.2}
\newcommand{\FrontierFourteenMinusTwoCILow}{\ensuremath{-}12.5}
\newcommand{\FrontierFourteenMinusTwoCIHigh}{1.2}
\newcommand{\FrontierThirtyMinusTwo}{\ensuremath{-}91.2}
\newcommand{\FrontierThirtyMinusTwoCILow}{\ensuremath{-}95.0}
\newcommand{\FrontierThirtyMinusTwoCIHigh}{\ensuremath{-}83.8}
\newcommand{\FrontierThirtyBudgetExceeded}{69}

\newcommand{\FrontierThirtyInfeasible}{9}
\newcommand{\FrontierThirtyAgentFailures}{11}
\newcommand{\FrontierThirtyMedianCalls}{20}
\newcommand{\FrontierThirtyCallsIQRLow}{18}
\newcommand{\FrontierThirtyCallsIQRHigh}{22}
\newcommand{\FrontierThirtyMaxCalls}{23}
\newcommand{\FrontierThirtyMedianTokens}{132{,}392}

\newcommand{\FrontierFourteenMedianPromptTokens}{79{,}898}

\newcommand{\RetrievalAssignedLoss}{79.6}
\newcommand{\RetrievalAssignedLossCILow}{60.4}
\newcommand{\RetrievalAssignedLossCIHigh}{94.2}
\newcommand{\RetrievalStressGain}{58.1}
\newcommand{\RetrievalStressGainCILow}{36.1}
\newcommand{\RetrievalStressGainCIHigh}{79.9}
\newcommand{\RetrievalSelectivity}{58.1}

\newcommand{\RetrievalHolmP}{.001}

\newcommand{\VisibleRepairAssignedLoss}{17.5}
\newcommand{\VisibleRepairAssignedLossCILow}{14.9}
\newcommand{\VisibleRepairAssignedLossCIHigh}{20.2}
\newcommand{\VisibleRepairStressGain}{4.9}
\newcommand{\VisibleRepairStressGainCILow}{2.6}
\newcommand{\VisibleRepairStressGainCIHigh}{7.0}
\newcommand{\VisibleRepairSelectivity}{4.9}
\newcommand{\VisibleRepairSelectivityCILow}{2.2}
\newcommand{\VisibleRepairSelectivityCIHigh}{7.5}

\newcommand{\VisibleRepairHolmP}{.003}

\newcommand{\WorkflowPromptAssignedLoss}{2.4}
\newcommand{\WorkflowPromptAssignedLossCILow}{\ensuremath{-}0.4}
\newcommand{\WorkflowPromptAssignedLossCIHigh}{5.0}
\newcommand{\WorkflowPromptStressGain}{1.7}
\newcommand{\WorkflowPromptStressGainCILow}{\ensuremath{-}0.6}
\newcommand{\WorkflowPromptStressGainCIHigh}{3.8}
\newcommand{\WorkflowPromptSelectivity}{0.0}
\newcommand{\WorkflowPromptSelectivityCILow}{\ensuremath{-}2.4}
\newcommand{\WorkflowPromptSelectivityCIHigh}{2.4}

\newcommand{\WorkflowPromptHolmP}{.949}

\newcommand{\InterfaceGuideAssignedLoss}{1.5}
\newcommand{\InterfaceGuideAssignedLossCILow}{\ensuremath{-}0.1}
\newcommand{\InterfaceGuideAssignedLossCIHigh}{3.1}
\newcommand{\InterfaceGuideStressGain}{0.1}
\newcommand{\InterfaceGuideStressGainCILow}{\ensuremath{-}1.5}
\newcommand{\InterfaceGuideStressGainCIHigh}{1.9}
\newcommand{\InterfaceGuideSelectivity}{0.7}
\newcommand{\InterfaceGuideSelectivityCILow}{\ensuremath{-}1.7}
\newcommand{\InterfaceGuideSelectivityCIHigh}{3.1}

\newcommand{\WorkflowPromptTOSTMargin}{3.6}
\newcommand{\InterfaceGuideTOSTMargin}{1.7}
\newcommand{\SentinelGroups}{480}

\newcommand{\SentinelPassOne}{89.7}
\newcommand{\SentinelPassAll}{82.7}
\newcommand{\SentinelPassAny}{94.8}
\newcommand{\SentinelDisagreement}{12.1}
\newcommand{\SentinelRoutedGain}{7.1}
\newcommand{\SentinelRoutedGainCILow}{2.5}
\newcommand{\SentinelRoutedGainCIHigh}{11.7}
\newcommand{\SentinelWorkflowPromptGain}{\ensuremath{-}1.2}
\newcommand{\SentinelWorkflowPromptGainCILow}{\ensuremath{-}2.5}
\newcommand{\SentinelWorkflowPromptGainCIHigh}{0.0}
\newcommand{\SentinelInterfaceGuideGain}{\ensuremath{-}0.4}
\newcommand{\SentinelInterfaceGuideGainCILow}{\ensuremath{-}3.3}
\newcommand{\SentinelInterfaceGuideGainCIHigh}{2.5}
\newcommand{\SentinelFamilies}{4}
\newcommand{\StructuralAssignedLoss}{96.1}
\newcommand{\StructuralAssignedLossCILow}{93.6}
\newcommand{\StructuralAssignedLossCIHigh}{98.0}
\newcommand{\StructuralStressGain}{96.6}
\newcommand{\StructuralStressGainCILow}{94.6}
\newcommand{\StructuralStressGainCIHigh}{98.5}
\newcommand{\StructuralSelectivity}{95.1}
\newcommand{\StructuralSelectivityCILow}{92.6}
\newcommand{\StructuralSelectivityCIHigh}{97.3}

\newcommand{\StructuralHolmP}{\ensuremath{<}.001}

\newcommand{\RegisteredStructuralStressGain}{95.8}
\newcommand{\RegisteredStructuralStressGainCILow}{89.6}
\newcommand{\RegisteredStructuralStressGainCIHigh}{100.0}

\newcommand{\RegisteredStructuralSpecificity}{93.8}
\newcommand{\RegisteredStructuralSpecificityCILow}{85.4}
\newcommand{\RegisteredStructuralSpecificityCIHigh}{102.1}

\newcommand{\CanonicalControlAssignedLoss}{\ensuremath{-}1.2}
\newcommand{\CanonicalControlAssignedLossCILow}{\ensuremath{-}3.7}
\newcommand{\CanonicalControlAssignedLossCIHigh}{1.2}
\newcommand{\CanonicalControlStressGain}{\ensuremath{-}0.2}
\newcommand{\CanonicalControlStressGainCILow}{\ensuremath{-}2.5}
\newcommand{\CanonicalControlStressGainCIHigh}{1.7}

\newcommand{\BaselineThirtyOverBudget}{389}

\newcommand{\BaselineThirtyRows}{408}
\newcommand{\BaselineThirtyCalls}{18.9}
\newcommand{\BaselineThirtyRequests}{19.8}
\newcommand{\BaselineThirtyTokens}{130{,}468}

\newcommand{\ControllerThirtyOverBudget}{3}
\newcommand{\ControllerThirtyCalls}{29.9}
\newcommand{\ControllerThirtyControllerCalls}{28.7}
\newcommand{\ControllerThirtyModelCalls}{1.2}
\newcommand{\ControllerThirtyRequests}{2.2}
\newcommand{\ControllerThirtyTokens}{21{,}159}
\newcommand{\BaselineTwoVTS}{96.7}
\newcommand{\ControllerTwoVTS}{97.4}
\newcommand{\StructuralSuccessSuperior}{17}
\newcommand{\StructuralPareto}{17}
\newcommand{\StructuralCostSuperior}{4}

\newcommand{\RetrievalSuccessSuperior}{21}
\newcommand{\RetrievalPareto}{21}
\newcommand{\RetrievalCostSuperior}{0}
\newcommand{\RetrievalDecisionN}{21}
\newcommand{\VisibleSuccessSuperior}{0}
\newcommand{\VisiblePareto}{9}
\newcommand{\VisibleCostSuperior}{1}
\newcommand{\VisibleDecisionN}{21}
\newcommand{\WorkflowSuccessSuperior}{0}
\newcommand{\WorkflowPareto}{10}
\newcommand{\WorkflowCostSuperior}{1}
\newcommand{\WorkflowDecisionN}{20}
\newcommand{\InterfaceSuccessSuperior}{0}
\newcommand{\InterfacePareto}{9}
\newcommand{\InterfaceCostSuperior}{1}
\newcommand{\InterfaceDecisionN}{20}
\newcommand{\LocalizationPackets}{127}
\newcommand{\LocalizationVisible}{87}
\newcommand{\LocalizationWorkflow}{28}
\newcommand{\LocalizationReturned}{12}
\newcommand{\LocalizationMultiExcluded}{36}
\newcommand{\LocalizationInfraExcluded}{2}

\newcommand{\BAConstant}{33.3}

\newcommand{\BAObligation}{53.4}
\newcommand{\BAObligationCILow}{44.6}
\newcommand{\BAObligationCIHigh}{60.1}
\newcommand{\BATermination}{48.5}
\newcommand{\BATerminationCILow}{43.0}
\newcommand{\BATerminationCIHigh}{55.4}
\newcommand{\BAJoint}{48.9}
\newcommand{\BAJointCILow}{42.4}
\newcommand{\BAJointCIHigh}{58.9}
\newcommand{\BATrace}{51.3}
\newcommand{\BATraceCILow}{44.8}
\newcommand{\BATraceCIHigh}{57.0}
\newcommand{\BACallsOnly}{54.8}
\newcommand{\BACallsOnlyCILow}{47.8}
\newcommand{\BACallsOnlyCIHigh}{61.3}
\newcommand{\BATraceOnly}{52.4}
\newcommand{\BATraceOnlyCILow}{44.2}
\newcommand{\BATraceOnlyCIHigh}{60.2}
\newcommand{\BAVerdictVector}{52.5}
\newcommand{\BAVerdictVectorCILow}{44.3}
\newcommand{\BAVerdictVectorCIHigh}{62.4}
\newcommand{\BAMenuSize}{71.8}
\newcommand{\BAMenuSizeCILow}{60.4}
\newcommand{\BAMenuSizeCIHigh}{82.6}
\newcommand{\BACorrected}{54.1}
\newcommand{\BACorrectedCILow}{45.9}
\newcommand{\BACorrectedCIHigh}{63.2}
\newcommand{\CorrectedReturnedRecall}{8.3}

\newcommand{\JointReturnedRecovered}{1}
\newcommand{\GainTraceVsCalls}{\ensuremath{-}3.5}
\newcommand{\GainTraceVsCallsLow}{\ensuremath{-}9.4}
\newcommand{\GainTraceVsCallsHigh}{5.1}
\newcommand{\GainTraceVsObligation}{\ensuremath{-}2.1}
\newcommand{\GainTraceVsObligationLow}{\ensuremath{-}15.1}
\newcommand{\GainTraceVsObligationHigh}{12.4}
\newcommand{\GainJointVsObligation}{\ensuremath{-}4.4}
\newcommand{\GainJointVsObligationLow}{\ensuremath{-}15.0}
\newcommand{\GainJointVsObligationHigh}{9.7}
\newcommand{\GainTraceVsTraceOnly}{\ensuremath{-}1.1}
\newcommand{\GainTraceVsTraceOnlyLow}{\ensuremath{-}7.6}
\newcommand{\GainTraceVsTraceOnlyHigh}{6.8}
\newcommand{\GainTraceVsJoint}{2.3}
\newcommand{\GainTraceVsJointLow}{\ensuremath{-}9.1}
\newcommand{\GainTraceVsJointHigh}{10.3}
\newcommand{\ChallengeRows}{83}

\newcommand{\ChallengeBA}{62.4}
\newcommand{\ChallengeReturnedCorrect}{3}
\newcommand{\ChallengeReturnedCases}{26}
\newcommand{\CrossoverConfigs}{4}

\newcommand{\CrossoverRowsRead}{6{,}912}
\newcommand{\CrossoverRowsKept}{6{,}904}
\newcommand{\CrossoverRowsExcluded}{8}
\newcommand{\CrossoverRepeats}{2}

\newcommand{\CrossoverCapHigh}{4{,}096}
\newcommand{\CrossoverManipulationCeiling}{5}

\newcommand{\CrossoverDisagreementCapLow}{24.2}
\newcommand{\CrossoverDisagreementCapHigh}{21.2}
\newcommand{\CrossoverBMenuCapLow}{\ensuremath{-}23.6}
\newcommand{\CrossoverBMenuCapLowCILow}{\ensuremath{-}28.5}
\newcommand{\CrossoverBMenuCapLowCIHigh}{\ensuremath{-}18.8}
\newcommand{\CrossoverTruncBCapLowMin}{28.8}
\newcommand{\CrossoverTruncBCapLowMax}{35.8}
\newcommand{\CrossoverBMenuCapHigh}{\ensuremath{-}21.3}
\newcommand{\CrossoverBMenuCapHighCILow}{\ensuremath{-}26.7}
\newcommand{\CrossoverBMenuCapHighCIHigh}{\ensuremath{-}16.1}
\newcommand{\CrossoverTruncBCapHighMin}{20.2}
\newcommand{\CrossoverTruncBCapHighMax}{25.3}
\newcommand{\CrossoverBInteraction}{2.1}

\newcommand{\CrossoverBInteractionFamilyCILow}{\ensuremath{-}2.6}
\newcommand{\CrossoverBInteractionFamilyCIHigh}{6.8}
\newcommand{\CrossoverBInteractionBoundLow}{1.9}
\newcommand{\CrossoverBInteractionBoundHigh}{2.6}
\newcommand{\CrossoverNInteraction}{\ensuremath{-}0.3}
\newcommand{\CrossoverNInteractionCILow}{\ensuremath{-}4.5}
\newcommand{\CrossoverNInteractionCIHigh}{3.8}
\newcommand{\CrossoverSInteraction}{1.4}
\newcommand{\CrossoverSInteractionCILow}{\ensuremath{-}8.3}
\newcommand{\CrossoverSInteractionCIHigh}{11.1}
\newcommand{\MenuAllToolsSuccess}{\ensuremath{-}2.9}
\newcommand{\MenuAllToolsSuccessCILow}{\ensuremath{-}7.0}
\newcommand{\MenuAllToolsSuccessCIHigh}{0.5}
\newcommand{\MenuAllToolsSimilarity}{\ensuremath{-}0.2}
\newcommand{\MenuAllToolsSimilarityCILow}{\ensuremath{-}5.5}
\newcommand{\MenuAllToolsSimilarityCIHigh}{4.5}
\newcommand{\MenuAllToolsTokens}{16{,}916}
\newcommand{\MenuAllToolsTokensCILow}{15{,}021}
\newcommand{\MenuAllToolsTokensCIHigh}{18{,}676}
\newcommand{\MenuAllToolsRequests}{0.86}
\newcommand{\MenuAllToolsRequestsCILow}{0.56}
\newcommand{\MenuAllToolsRequestsCIHigh}{1.18}

\newcommand{\MenuAllToolsSuccessP}{.224}

\newcommand{\MenuTenSuccess}{\ensuremath{-}0.5}
\newcommand{\MenuTenSuccessCILow}{\ensuremath{-}1.8}
\newcommand{\MenuTenSuccessCIHigh}{0.8}
\newcommand{\MenuTenSimilarity}{1.6}
\newcommand{\MenuTenSimilarityCILow}{\ensuremath{-}0.8}
\newcommand{\MenuTenSimilarityCIHigh}{4.0}
\newcommand{\MenuTenTokens}{5{,}322}
\newcommand{\MenuTenTokensCILow}{4{,}622}
\newcommand{\MenuTenTokensCIHigh}{6{,}018}

\newcommand{\MenuTenDuration}{1.07}
\newcommand{\MenuTenDurationCILow}{0.28}
\newcommand{\MenuTenDurationCIHigh}{1.85}

\newcommand{\MenuScenarios}{32}
\newcommand{\HistoricalMenuStressContrast}{\ensuremath{-}0.7}
\newcommand{\HistoricalMenuStressContrastCILow}{\ensuremath{-}5.2}
\newcommand{\HistoricalMenuStressContrastCIHigh}{3.9}
\newcommand{\ExternalRho}{\ensuremath{-}0.8}
\newcommand{\ExternalRhoCILow}{\ensuremath{-}1.0}
\newcommand{\ExternalRhoCIHigh}{\ensuremath{-}0.5}
\newcommand{\ExternalRhoConfigCILow}{\ensuremath{-}1.0}
\newcommand{\ExternalRhoConfigCIHigh}{\ensuremath{-}0.11}
\newcommand{\ExternalKendall}{\ensuremath{-}0.6}

\newcommand{\PilotBlocks}{192}
\newcommand{\PilotIntents}{12}
\newcommand{\PilotEpisodes}{1{,}536}

\newcommand{\PilotBundle}{81.8}
\newcommand{\PilotBundleBlocks}{157}
\newcommand{\PilotRouter}{78.6}
\newcommand{\PilotRouterBlocks}{151}
\newcommand{\PilotRetry}{67.7}
\newcommand{\PilotRetryBlocks}{130}
\newcommand{\PilotSwitch}{99.0}
\newcommand{\PilotSwitchBlocks}{190}
\newcommand{\PilotTrace}{97.9}
\newcommand{\PilotTraceBlocks}{188}

\newcommand{\PilotTraceDifference}{\ensuremath{-}1.0}
\newcommand{\PilotTraceDifferenceLow}{\ensuremath{-}2.6}
\newcommand{\PilotTraceDifferenceHigh}{0.0}
\newcommand{\PilotTraceCostPerThousand}{4.49}
\newcommand{\PilotSwitchCostPerThousand}{4.62}
\newcommand{\PilotSwitchNoProbeCostPerThousand}{1.94}
\newcommand{\TotalCost}{768.38}

\newcommand{\CostMostExpensiveModel}{Claude Opus 5}
\newcommand{\CostMostExpensiveUSD}{165.18}
\newcommand{\CostCheapestModel}{Qwen3.7 Flash}
\newcommand{\CostCheapestUSD}{0.88}

\newcommand{\RecomputedAnalyses}{16}
\newcommand{\PreservedAnalyses}{3}

\newcommand{\TestsPassed}{242}
\newcommand{\SubtestsPassed}{227}
\newcommand{\JUnitCases}{469}
\newcommand{\CertificateModels}{8}

\newcommand{\TruncatedFirstConfigs}{9}
\newcommand{\TruncatedFirstTotal}{171}
\newcommand{\TruncatedFirstReasoningOnly}{120}
\newcommand{\TruncatedFirstReasoningOnlyShare}{70.2}
\newcommand{\TruncatedFirstPartialCall}{39}
\newcommand{\TruncatedFirstPartialText}{12}
\newcommand{\TruncatedFirstZeroConfigs}{5}
\newcommand{\TruncatedFirstRequestsPerConfig}{240}
\newcommand{\SweepCapLow}{1{,}024}
\newcommand{\SweepCapHigh}{16{,}384}
\newcommand{\SweepConfigs}{8}

\newcommand{\SweepBPassed}{6}
\newcommand{\SweepRowsKept}{13{,}824}
\newcommand{\SweepRowsRead}{13{,}824}

\newcommand{\SweepRepeats}{2}

\newcommand{\SweepBTruncLowMin}{30.6}
\newcommand{\SweepBTruncLowMax}{69.8}

\newcommand{\SweepBTruncHighMax}{1.4}

\newcommand{\SweepDisagreementLow}{17.5}
\newcommand{\SweepDisagreementHigh}{8.1}
\newcommand{\SweepBPassedMenuLow}{\ensuremath{-}43.7}
\newcommand{\SweepBPassedMenuLowCILow}{\ensuremath{-}48.5}
\newcommand{\SweepBPassedMenuLowCIHigh}{\ensuremath{-}39.1}
\newcommand{\SweepBPassedMenuHigh}{\ensuremath{-}2.7}
\newcommand{\SweepBPassedMenuHighCILow}{\ensuremath{-}4.7}
\newcommand{\SweepBPassedMenuHighCIHigh}{\ensuremath{-}0.9}
\newcommand{\SweepBPassedInteraction}{41.0}
\newcommand{\SweepBPassedInteractionCILow}{36.2}
\newcommand{\SweepBPassedInteractionCIHigh}{45.5}
\newcommand{\SweepMMenuLow}{\ensuremath{-}2.4}
\newcommand{\SweepMMenuHigh}{0.3}
\newcommand{\SweepMInteraction}{2.8}
\newcommand{\SweepMInteractionCILow}{\ensuremath{-}3.1}
\newcommand{\SweepMInteractionCIHigh}{8.0}
\newcommand{\SweepNMenuLow}{\ensuremath{-}3.1}
\newcommand{\SweepNMenuHigh}{\ensuremath{-}2.1}
\newcommand{\SweepNInteraction}{1.0}
\newcommand{\SweepNInteractionCILow}{\ensuremath{-}2.4}
\newcommand{\SweepNInteractionCIHigh}{4.2}

\newcommand{\SweepBDriftMin}{\ensuremath{-}11.8}
\newcommand{\SweepBDriftMax}{\ensuremath{-}5.2}
\newcommand{\SweepClaudeHaikuFiveFiveTruncLow}{4.5}

\newcommand{\SweepGPTSixLunaTruncLow}{0.0}

\newcommand{\PilotFailedProbes}{65}
\newcommand{\PilotSuccessfulProbes}{127}
\newcommand{\PilotRouterRecovered}{40}
\newcommand{\PilotRouterRecoveredPct}{61.5}
\newcommand{\PilotBundleRecovered}{43}
\newcommand{\PilotBundleRecoveredPct}{66.2}
\newcommand{\PilotRetryRecovered}{16}
\newcommand{\PilotRetryRecoveredPct}{24.6}
\newcommand{\PilotSwitchRecovered}{64}
\newcommand{\PilotSwitchRecoveredPct}{98.5}
\newcommand{\PilotBudgetRecoveredPct}{41.5}
\newcommand{\PilotBundlePolicyPct}{88.5}
\newcommand{\PilotSwitchPolicyPct}{99.5}

\newcommand{\ConfigGeminiThreeSixFlashWorkflowEffect}{\ensuremath{-}30.6}

\newcommand{\ConfigHYFourPreviewWorkflowEffect}{34.7}

\title{ToolBottleneckBench: Measuring Tool-Agent Performance\\ under Controlled Pressures and Fixed Harness Budgets}

\providecommand{\BottleneckAuthorBlock}{Anonymous ACL submission}
\providecommand{\BottleneckPdfAuthor}{}
\author{\BottleneckAuthorBlock}

\begin{document}
\maketitle

\hypersetup{
  pdftitle={ToolBottleneckBench: Measuring Tool-Agent Performance under Controlled Pressures and Fixed Harness Budgets},
  pdfauthor={\BottleneckPdfAuthor}}

\begin{abstract}
The success score of a tool agent reflects both its model and the budgets of the harness that runs it.
We present ToolBottleneckBench, a benchmark that measures tool-agent performance under controlled pressures and fixed harness budgets. Synthetic enterprise tasks stay fixed while one pressure varies: the registered tool catalog, the visible tool menu, the certified workflow, or the encoding of returned records. An exact grader records the first unmet obligation of each failed episode and the harness limit, if any, that ended it.
Across \CoreModels{} model and harness configurations, the visible menu is the only one of the three pressures that act on the model with a large penalty, and this penalty tracks truncation at the completion cap. Raising the cap sixteenfold removes most of the penalty in the truncation-heavy configurations. Most failures end at a harness limit, and the failure record identifies the assigned pressure no more reliably than a count of tool calls. The repairs that help change which tools reach the model or who issues the calls, and only two repairs, neither improving model behavior, beat every configuration switch.
We recommend reporting pressure effects with the harness budgets under which they were measured. We release the benchmark at \ifdefined\BottleneckPreprint\url{https://github.com/zichuanxu/tool-bottleneck-bench}\else\url{https://anonymous.4open.science/r/TBB-0307}\fi.
\end{abstract}
\section{Introduction}
\label{sec:intro}

Tool agents are usually compared by a single success score, yet an agent runs
inside a harness that shows it a menu of tools, returns tool results in
some format, and enforces budgets on tokens, calls, and turns. When an agent
fails, the cause may lie in the model or in these limits, and neither a success
score nor a label for the first failed step can tell the two apart. An agent
facing a large menu whose first response is cut off by the per-request
completion cap fails before it has chosen any tool, yet a grader that checks
obligations in order records a tool-selection failure. A serving adapter
that silently drops well-formed tool calls has a similar effect
\citep{wang2026trajectorycensoring}. Recent benchmarks report results per model
and harness \citep{yao2026harnessbench,kapoor2026hal} but do not measure how
much of a load's penalty occurs together with a harness limit.

Realistic environments change several
loads at once \citep{lu2025toolsandbox,yao2025taubench,trivedi2024appworld}.
Controlled benchmarks vary loads with the goal fixed
\citep{babu2026toolmenubench,maekawa2026funcbenchgen} but do not record which
harness limit ended the failing episodes. Failure diagnosis localizes the
failing stage \citep{raj2026modelorharness} but, without
an assigned pressure, cannot check whether the label is right.

\begin{figure*}[t]
  \centering
  \includegraphics[width=\textwidth]{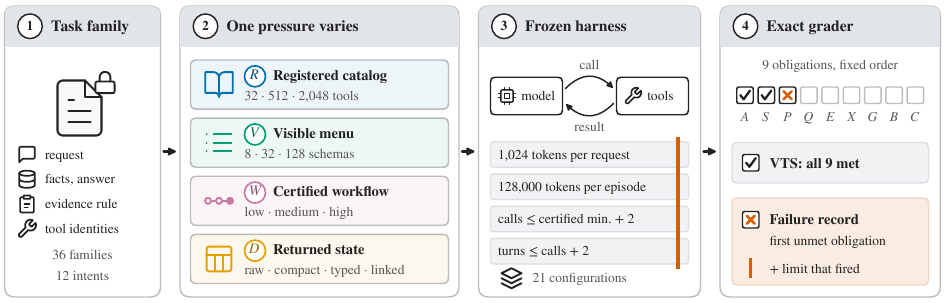}
  \caption{Overview. (1) A task family fixes the request, facts, answer,
  evidence rule, and tools. (2) Its matched variants differ in one assigned
  pressure ($R$, $V$, $W$, or $D$). (3) All main results use the same frozen
  budgets, and (4) the exact grader records the first unmet of nine obligations
  and the harness limit, if any, that fired. $R$ is scored by intention-to-treat (ITT) success, which
  counts episodes whose tools were never retrieved as failures, so its scale
  differs from that of $V$, $W$, and $D$.}
  \label{fig:concept}
  \Description{A schematic with four panels from left to right: a task family panel with a locked document and the elements that stay fixed, four pressure chips R, V, W, and D with a pictogram and the levels of each, the frozen harness contract as a model-tools loop above four limit bars, and the exact grader as a row of nine obligation checkboxes with the first unmet one crossed, followed by the two outcomes, VTS and the failure record.}
\end{figure*}

We introduce ToolBottleneckBench, a benchmark that measures tool-agent performance under controlled pressures and fixed harness budgets (\figref{fig:concept}). A task family fixes the request, the facts, the answer, the evidence rule, and the tools. A deterministic compiler then changes one pressure at a time to produce \emph{matched variants}. 
The four pressures are the registered tool catalog ($R$), the visible tool menu ($V$), the certified workflow ($W$), and the encoding of returned records ($D$). 
All main results use one frozen harness contract. An exact grader checks nine obligations and records the first unmet one together with a terminal flag that names the harness limit, if any, that fired. We call this pair the \emph{failure record}. 

With this design we ask four questions.
\begin{inparaenum} [(1)]
  \item Which pressures reduce verified task success (VTS)?
  \item How much of each loss occurs together with a harness limit?
  \item Where do failed episodes end, and does the failure record reveal which pressure was assigned?
  \item Which repairs help under their pressure, and do they beat switching to another configuration?
\end{inparaenum}

We evaluate \CoreModels{} configurations from \Organizations{} organizations on
\TotalRows{} scored conditions. Our main findings are as follows.
\begin{inparaenum} [(1)]
\item Of the three model-facing pressures ($V$, $W$, $D$), only the visible menu has a large penalty. Growing it from 8 to 128 tool schemas changes VTS by \VisibleEffect{} points under a routed menu and by $-$\VisibleRepairAssignedLoss{} points when the full menu stays visible. The workflow pressure has no detectable pooled effect but opposite-signed effects across configurations, and returned-state encoding has small effects.
\item Under the routed menu, the menu penalty tracks completion-cap truncation across configurations (Pearson $r=\VisibleTruncationPearson$), the \ZeroTruncationConfigs{} configurations that never truncate stay within \ZeroTruncationEffectMax{} points of zero, and in the \SweepBPassed{} truncation-heavy configurations a sixteenfold cap removes most of the menu penalty, from \SweepBPassedMenuLow{} to \SweepBPassedMenuHigh{} points.
\item Of all exactly graded failures, \HarnessLimitShare{}\% end at a harness limit and only \WrongConclusionShare{}\% are wrong conclusions. Our decoders identify the assigned pressure from the failure record no more reliably than from a count of tool calls.
\item The repairs that help change which tools reach the model or who issues the calls, and our prompt repairs do not. Only two repairs, neither improving model behavior, beat every configuration switch on success.
\end{inparaenum}

We contribute the benchmark, its failure record, and a repair protocol that tests each repair with and without its pressure and against every configuration switch, and release the generator, grader, scored rows, and analysis code. %
\section{Related Work}
\label{sec:related}

\paragraph{Tool-use benchmarks.}
Early benchmarks score whether a model calls APIs correctly
\citep{li2023apibank,qin2024toolllm,patil2025bfcl}. Later ones use stateful,
realistic environments \citep{lu2025toolsandbox,yao2025taubench,trivedi2024appworld}
or live MCP servers \citep{wang2025mcpbench,luo2025mcpuniverse}, with
StableToolBench stabilizing unstable APIs \citep{guo2024stabletoolbench}.
In these environments, two tasks usually differ in catalog, workflow, and
observation format at once, so a score difference cannot be traced to one load.

\paragraph{Controlled studies of tool load.}
Other work studies specific loads with the goal fixed: tool retrieval
\citep{shi2025toolret,liu2026toolscope}, connector-first routing
\citep{fei2025mcpzero,gan2025ragmcp}, menu size
\citep{repantis2026howmanytools}, distractors and dependency depth
\citep{maekawa2026funcbenchgen}, long API chains \citep{wan2026apiflowbench},
and catalog size or output length \citep{kate2025longfunceval,kate2026tooloutputs},
with related position effects in long contexts \citep{liu2024lostmiddle} and a
loss from context length alone \citep{du2025contextlength}.
ToolMenuBench separates whether the gold tool is visible from whether it is
selected \citep{babu2026toolmenubench,babu2026cmtf}, and we follow this
distinction. ToolBottleneckBench varies catalog, menu, workflow, and encoding on the same
task families and records which harness limit ended each failing episode.

\paragraph{Harness effects.}
Harness-Bench and HAL report results per model and harness
\citep{yao2026harnessbench,kapoor2026hal}. A harness change can reorder models
\citep{zhang2026discloseharness,li2026rightfit}, help one model while hurting
another \citep{cho2026nonmonotoneharness,starace2026scaffoldgaia}, or flip as
many tasks as a rerun \citep{liu2026harnessbuy}. The agent-computer interface
\citep{yang2024sweagent,wang2026aftbench}, system components
\citep{lukassen2026cafe}, context growth and pruning
\citep{zeng2026locabench,lodha2026lesscontext}, working memory
\citep{chen2026measurememory}, and serving adapters
\citep{wang2026trajectorycensoring} also change outcomes, as does deliberation
that displaces action \citep{cuadron2025overthinking}. We instead measure how
much of a controlled pressure effect occurs together with a specific harness
limit.

\paragraph{Failure diagnosis and repair.}
We build on stage attribution, counterfactual correction, and wrong-layer repair
controls \citep{zheng2026toolrobustbench,shams2026layerragbench,ma2026dover}.
Prior studies ask whether a failure belongs to the model or the harness
\citep{raj2026modelorharness}, note that localizing an error does not say what
to change \citep{yuan2026daaf}, patch harness flaws \citep{chen2026harnessfix},
and show that readout intervention beats prompting
\citep{chen2026lookingnotpicking}, that patches can be validated by replay
\citep{lin2026reflect} or still hurt \citep{yoon2026diagnosisnotprescription},
and that re-execution alone repairs some failures
\citep{luan2026repairorresample}. Exact graders need validity audits of their own
\citep{bhat2026benchmarkingbenchmarks,gurram2026agentprop,zhao2026calibration}.
Appendix~\ref{app:obligations} reports ours. ToolBottleneckBench combines these
techniques on one population with assigned pressures and compares every repair
with configuration switches, and Appendix~\ref{app:neighbors} compares designs.
\section{ToolBottleneckBench}
\label{sec:bench}

ToolBottleneckBench places a known bottleneck on a task, measures how
performance changes, and checks whether the recorded failure points to it. We
call such a controlled bottleneck an \emph{assigned pressure}. 

\subsection{Task Families and Matched Variants}
\label{sec:variants}

A \emph{task family} fixes everything that defines the correct outcome of a
task: the user request, the fact values that decide it, the decision label, the
rule for which evidence must be cited, the permitted final state of the
environment, and the identities of the tools that do the work (the
operational tools). The benchmark has \SemanticFamilies{} families
nested in \Intents{} synthetic enterprise intents over accounting, email, file,
and calendar records. Record IDs are opaque, so no answer can be reused across
tasks.

A deterministic compiler takes a family and changes one pressure to produce
matched variants. Matched variants differ only in that pressure. Under the workflow pressure,
the required \texttt{sources} array in the answer also grows from one to three
entries. Because the decision and facts of the correct answer are the same
across variants, the paired difference in success between matched variants
measures the effect of the pressure that was changed.

For example, one family asks whether an overdue case should be escalated. Its
decision facts (status, commitment date, and contact), its six operational tools,
and its reference answer, \texttt{\{"decision":"escalate", \ldots,
"sources":["source-2"]\}}, are fixed. Its $V$ variants differ only in how
many tool schemas are shown, and its $D$ variants only in how the same records
are encoded (\figref{fig:variants} in Appendix~\ref{app:protocol}).

\subsection{Four Assigned Pressures}
\label{sec:pressures}

Table~\ref{tab:pressures} summarizes the four pressures, or layers,
and what each one changes besides the intended load.

\begin{table*}[t]
\centering
\footnotesize
\caption{The four assigned pressures, each varied with the task family fixed.
Calls and schema bytes are means over the routed capacity episodes (Section~\ref{sec:harness}), prompt tokens over the fixed-menu lane, and observation bytes over the reference
programs.}
\label{tab:pressures}
\begin{tabular}{@{}l l p{0.52\linewidth}@{}}
\toprule
Layer & Levels & What the variant changes (measured) \\
\midrule
$R$ catalog & 32, 512, 2{,}048 tools & catalog the retriever searches; required-tool availability at $R=2{,}048$: flat 19.4\%, connector-first 77.8\% \\
$V$ menu & 8, 32, 128 tools & schema bytes per exposure 3{,}481 / 13{,}820 / 55{,}176; mean prompt tokens per episode (fixed menu) 15{,}714 at $V=8$ and 37{,}659 at $V=128$ \\
$W$ workflow & low, medium, high & certified program length; mean attempted calls 3.7 / 5.8 / 7.2 \\
$D$ returned state & raw, compact, typed, linked & encoding of returned records; mean observation bytes 1{,}214 / 1{,}471 / 1{,}351 / 1{,}533 \\
\bottomrule
\end{tabular}

\end{table*}

\paragraph{Registered catalog $R$.}
The catalog is the set of functions the system has registered. Nested catalogs
hold $R\in\{32,512,2{,}048\}$ functions. A frozen retriever, which does not depend
on the model, returns a 16-function shortlist from the catalog. We use two
retrievers: a flat lexical scorer and a connector-first router that first
picks a connector and then ranks its tools. We record separately whether the
required tools are in the shortlist (availability $A$) and whether the model
selects them when they are (selection $S$). Success on this pressure is scored
intention-to-treat as $A\cdot S$ (ITT success), so an episode whose
required tools were never retrieved counts as a failure. The catalog has its own block, Discovery.

\paragraph{Visible menu $V$.}
The menu is the set of tool schemas shown to the model. $V\in\{8,32,128\}$
counts these schemas, and the operational tools are always among them. The
added functions are route neighbors that differ from an operational tool in
one of workspace, region, or collection. An offline audit executed the added
functions \DistractorExecutions{} times, and no call returned a record. The system
prompt states that the operational tools match the route string in the request,
so a larger menu adds candidates and context without ambiguity about which tools
apply: schema bytes per exposure grow from \SchemaBytesVEight{} to
\SchemaBytesVOneTwoEight{}.

\paragraph{Certified workflow $W$.}
We call a tool program certified when an independent checker proves that
it is minimal and that its calls respect the data dependencies. Low, medium, and
high $W$ require 2, 4, and 6 reads, plus one call when the task requires a
reversible action. The number of required calls, the dependency depth, the
spread of evidence, and the accumulated context grow together, so $W$
is a composite pressure. A separate
frontier block extends $W$ further. It places the same facts behind 0, 2,
6, or 14 public handoffs, records whose only content points to the next
record, for certified minima of 2, 6, 14, or 30 calls.

\paragraph{Returned-state encoding $D$.}
Tool results return the same records in one of four encodings: raw JSON,
dictionary-coded tuples with a visible field dictionary (compact), typed
objects with a type dictionary (typed), or typed objects with identity
links and a relation dictionary (linked). All four decode to the same
values, and compact is larger than raw in bytes (Appendix~\ref{app:protocol}).

The \emph{capacity factorial}
crosses the endpoints of the three model-facing
pressures, $\{8,128\}\times\{\mathrm{low},\mathrm{high}\}\times\{\mathrm{raw},
\mathrm{linked}\}$, and adds four checkpoint cells ($V=32$, medium $W$, compact,
and typed), each varying one pressure from the baseline cell
$(8,\mathrm{low},\mathrm{raw})$.

\subsection{Harness Contract}
\label{sec:harness}

All main results use one frozen harness contract. 
The completion cap limits the tokens the
provider may generate per request to \CompletionTokenCap{}, and a response cut
off at the cap ends the episode. The cap fits one complete tool call or
answer: across the \SuccessResponseEpisodes{} successful episodes, completion 
tokens per response have median \SuccessResponseTokensMedian{} and 99th percentile \SuccessResponseTokensPNN{}. Where reasoning tokens count against the cap, it also 
covers reasoning and is tight for truncation-heavy configurations, which the 
sixteenfold sweep tests (Appendix~\ref{app:capsweep}). 
The episode budget is \EpisodeTokenBudget{} reported tokens. The \emph{call
limit} is the certified minimum plus two, and the turn limit is the call
limit plus two. An episode stops early once success is impossible within the call limit (an infeasibility stop). These four limits are the \emph{harness budgets}, and a \emph{harness limit} fires when an episode ends at one of them or at an infeasibility stop.

The main capacity runs use a \emph{routed menu}: the model sees the full menu
until its first executed call to an operational tool, and the menu then narrows
to the operational tools. This narrowing uses the ground truth, so the routed
menu is an oracle-assisted reference. A paired fixed-menu lane keeps the
full menu throughout and is the no-repair arm of the routed menu
(Section~\ref{sec:repair-design}). Two further studies (Appendix~\ref{app:crossover}) and two pilot actions (Appendix~\ref{app:pilot}) raise the completion cap. All other results use the frozen budgets.

\subsection{Verified Task Success and Failure Record}
\label{sec:grading}

We measure performance as verified task success (VTS). An episode succeeds only
if it meets nine obligations, and VTS is their conjunction,
\[
Y=A\land S\land P\land Q\land E\land X\land G\land B\land C,
\]
checked in this order. $A$ is availability of the required tools, $S$ selection
of the operational tools, $P$ completion of a dependency-valid program
(workflow), $Q$ an exact JSON answer (answer), $E$ the exact set of
cited records (evidence), $X$ the exact permitted change of state
(safety), $G$ no unrequested change of state (governance), $B$ the episode budget and call limit, and $C$ completion integrity. Any dependency-valid call order is accepted, and no model
judge enters the score. The state checks for $X$ and $G$ follow AppWorld and
$\tau$-Bench \citep{trivedi2024appworld,yao2025taubench}.

Each failed episode receives one \emph{terminal flag}, which records how the episode ended. 
In order of precedence, the flags are (i) truncation at the completion cap, (ii) exceeding the episode budget or the call limit, (iii) reaching the turn limit, (iv) an early stop because success had become impossible within the call limit, (v) a wrong conclusion, and (vi) a break of another term of the contract. 
We call the pair of the first unmet obligation and the terminal flag the failure record. 
Appendix~\ref{app:obligations} gives the grading rules and the audits that check the generator and the grader.
\section{Experimental Setup}
\label{sec:setup}

\subsection{Configurations and Data}
\label{sec:panel}

The unit of evaluation is a \emph{configuration}: a served model together with its provider route and the harness settings (tool protocol, prompt template, transcript handling, retry policy, and budgets). 
We evaluate \CoreModels{} configurations from \Organizations{} organizations, all through managed inference endpoints.
Appendix~\ref{app:panel} lists their baseline success, per-configuration effects, and cost. 
In total there are \TotalRows{} scored conditions, of which \ExactVTSRows{} are graded by exact VTS. Qwen3.8 Flash has no prompt-repair, sentinel-repeat, or frontier rows, so those analyses use the other \AuxModels{} configurations.

\subsection{Contrasts and Statistical Analysis}
\label{sec:stats}

For the capacity factorial we report eight contrasts, each a paired
within-family difference in VTS in percentage points (pp): the main effects of
$V$ (128 minus 8), $W$ (high minus low), and linked encoding (linked minus raw),
their three pairwise interactions, and compact minus raw and typed minus raw at
the baseline cell. Each contrast averages \ContrastPairs{} family-by-configuration
differences. Our primary intervals are unadjusted 95\% percentile cluster bootstraps with
\BootstrapDraws{} draws that resample the \SemanticFamilies{} families. We repeat
the analysis with the \Intents{} intents and with the \CoreModels{}
configurations as the resampling unit, and correct each set of eight $p$ values
with the Holm procedure \citep{holm1979simple}. Discovery contrasts resample
\DiscoveryClusters{} clusters of required tools. The \InfrastructureConditions{}
conditions that failed for infrastructure reasons are kept and counted as
failures of their configuration. Appendix~\ref{app:capacity} gives the contrast
weights and the sensitivity analyses.

\subsection{Repairs and the Switch Comparison}
\label{sec:repair-design}

We test one repair aimed at each pressure: the connector-first router for $R$, the routed menu for $V$, a workflow prompt for $W$, and an interface guide for $D$. 
The two prompt repairs edit the system prompt (Appendix~\ref{app:protocol}). 
We also test a public-handoff controller, a deterministic program that follows the
public handoffs and issues the calls itself. 
Its evaluation, the structural holdout, uses the 2- and 30-call frontier cells over \HoldoutIntents{} held-out intents, distinct from the \Intents{} core intents, on \StructuralModels{} configurations.

A repair should help more under its pressure than without it. Let $\mu_{\ell z}^{a}$ be the mean success (VTS, or ITT success for $R$) for layer $\ell$ at pressure level $z\in\{0,1\}$ under intervention $a$, where $a=0$ means no repair and $a=\ell$ the repair for $\ell$, and let $\mathcal K_\ell$ be the repairs for other layers run on the same stressed cells.
We compute four quantities:
\begin{align*}
L_\ell&=\mu_{\ell0}^{0}-\mu_{\ell1}^{0},\qquad
\Gamma_\ell=\mu_{\ell1}^{\ell}-\mu_{\ell1}^{0},\\
\Sigma_\ell&=\Gamma_\ell-\bigl(\mu_{\ell0}^{\ell}-\mu_{\ell0}^{0}\bigr),\\
\Sigma_\ell^{k}&=\bigl(\mu_{\ell1}^{k}-\mu_{\ell1}^{0}\bigr)-\bigl(\mu_{\ell0}^{k}-\mu_{\ell0}^{0}\bigr),\\
\Delta_\ell&=\Sigma_\ell-\max_{k\in\mathcal K_\ell}\Sigma_\ell^{k}.
\end{align*}
$L$ is the assigned loss under the pressure, $\Gamma$ the stress gain of the repair under pressure, $\Sigma$ its selectivity, i.e., how much more it helps under pressure than without it, and $\Delta$ its wrong-layer advantage, its selectivity minus that of the best repair for another layer on the same cells.
We say a repair \emph{activates} if $L$, $\Gamma$, and $\Sigma$ are all positive, tested jointly with an intersection-union test \citep{berger1982multiparameter} and Holm correction over the four layers. 
The controller must also have $\Delta>0$ and is Holm-corrected together with its wrong-layer control (Appendix~\ref{app:repairs}).

A practitioner whose configuration fails can also switch to another
configuration, i.e., change the served model and its route while the harness
stays fixed. We re-execute each repair and every complete
configuration switch on the same stressed cells, rather than replaying saved
traces \citep{gonuguntla2026replaygap}, and compare them for each incumbent
configuration, in the spirit of cost-aware agent evaluation
\citep{kapoor2024agentsmatter}. A repair is success-superior or cost-superior
if its paired 95\% interval against every switch lies above zero on success or
below zero on cost, and Pareto-nondominated if no switch is at least as
good on success, tokens, wall time, and cost and strictly better on one.
\section{Results}
\label{sec:results}

\subsection{Which Pressures Reduce Verified Task Success?}
\label{sec:q-cost}

\begin{figure*}[t]
\centering
\includegraphics[width=\textwidth]{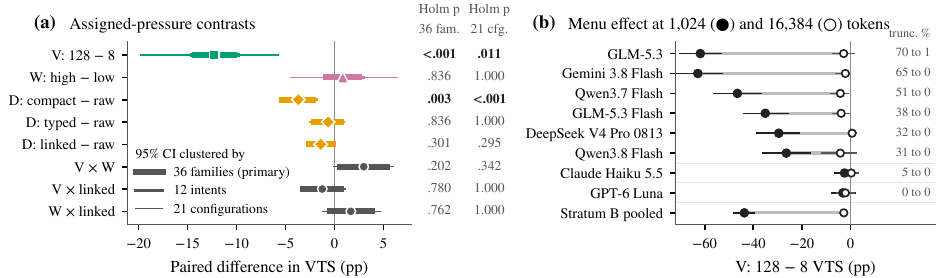}
\caption{The visible-menu penalty is large, and in truncation-heavy
configurations a sixteenfold completion cap removes most of it. (a) The eight contrasts as paired within-family differences in VTS
(pp), with nested 95\% intervals clustered by \SemanticFamilies{} families (thick,
primary), \Intents{} intents, and \CoreModels{} configurations (thin). (b) The sixteenfold cap
sweep: each configuration's visible-menu contrast at \SweepCapLow{} (filled) and
\SweepCapHigh{} (hollow) tokens with 95\% family-cluster intervals, its $V=128$
truncation shares, and the pooled contrast of the \SweepBPassed{} truncation-heavy
configurations (``Stratum B pooled'').}
\label{fig:capacity}
\Description{Panel a is a forest plot of eight contrasts with three interval
styles per row and a vertical zero line; only the visible-menu row lies far left
of zero. Panel b lists eight sweep configurations and a pooled row; each row
joins a filled point at the low cap, far left of zero for the six
truncation-heavy configurations, to a hollow point at the high cap near zero,
with truncation shares falling from tens of percent to about zero, while the two
control rows barely move.}
\end{figure*}

Mean VTS over the capacity episodes is \CoreVTS{}\%. Of the three pressures that
act on the model, only the visible menu has a large and robust effect
(\figref{fig:capacity}a). Growing the menu from 8 to 128 schemas changes VTS by
\VisibleEffect{}~pp $[\VisibleEffectCILow, \VisibleEffectCIHigh]$ (Holm
$p$~\VisibleHolmP{}), and \VisibleDegradedFamilies{} of the
\SemanticFamilies{} families degrade. The effect survives resampling configurations
($[\VisibleConfigCILow, \VisibleConfigCIHigh]$, Holm $p$~\HolmEq{\VisibleConfigHolmP})
or intents instead of families. With the full menu kept visible, the contrast is
$-$\VisibleRepairAssignedLoss{}~pp
$[-\VisibleRepairAssignedLossCIHigh, -\VisibleRepairAssignedLossCILow]$. Raising the workflow
pressure from low to high changes VTS by only \WorkflowEffectSigned{}~pp
$[\WorkflowEffectCILow, \WorkflowEffectCIHigh]$, so we detect no pooled workflow
effect within the call limit. The average hides opposite-signed configuration
effects, from \ConfigGeminiThreeSixFlashWorkflowEffect{}~pp (Gemini 3.6 Flash)
to $+$\ConfigHYFourPreviewWorkflowEffect{}~pp (HY4 Preview), with the three
largest gains in low-baseline configurations (Table~\ref{tab:panel} in Appendix~\ref{app:panel}).

The returned-state encoding has small effects. Compact encoding changes VTS by
\CompactEffect{}~pp $[\CompactEffectCILow, \CompactEffectCIHigh]$ (Holm
$p$~\HolmEq{\CompactHolmP}), but the exact sign-flip test over the
\Intents{} intents does not support it after correction (Holm
$p$~\HolmEq{\CompactExactIntentHolmP}). Typed and linked encodings show no
detectable effect. The $V\times W$ interaction, \InteractionEffectSigned{}~pp
$[\InteractionEffectCILow, \InteractionEffectCIHigh]$, does not survive
correction (Holm $p$~\HolmEq{\InteractionHolmP}, Appendix~\ref{app:capacity}). In ToolSandbox, a larger menu costs tokens more clearly than success (Appendix~\ref{app:external}).

\subsection{How Much of the Menu Penalty Occurs with the Completion Cap?}
\label{sec:q-budget}

Most of the menu penalty occurs with completion-cap truncation. Under the frozen
cap, the share of capacity episodes truncated at the cap rises from
\TruncationRateVEight{}\% at $V=8$ to \TruncationRateVOneTwoEight{}\% at $V=128$,
and \TruncatedBeforeFirstCallVOneTwoEight{} of the \TruncatedFailuresVOneTwoEight{}
truncated episodes at $V=128$ end before the model makes any call. For the
\TruncatedFirstConfigs{} configurations with reasoning-token telemetry,
\TruncatedFirstReasoningOnlyShare{}\% of truncated first responses to a
128-schema request hold reasoning tokens only, most of them from Gemini 3.8
Flash (Table~\ref{tab:truncated-first} in Appendix~\ref{app:crossover}). Of the
\VisiblePairs{} matched pairs, \PairsLost{} lose success at $V=128$ and
\PairsGained{} gain it. The \PairsLostTruncated{} lost pairs whose $V=128$
variant was truncated carry \PairsLostTruncatedPoints{} of the
\PairsLostPoints{}~pp lost.

Truncation depends on the assigned menu size, so we stratify on it
\citep{frangakis2002principal}: the \NonTruncatedPairs{} pairs that neither menu
truncated form a stratum the paired design observes directly, without trimming
\citep{lee2009bounds}, and within it the menu effect is
\NonTruncatedPairedEffect{}~pp. If every non-truncated episode keeps its outcome
and each truncated failure could have gone either way, worst-case bounds place
the total menu effect in $[\VisibleBoundLower, \VisibleBoundUpper]$~pp.

Across configurations, the $V$ effect tracks the truncation rate at $V=128$
(Pearson $r=\VisibleTruncationPearson$, Spearman
$\rho=\VisibleTruncationSpearman$, \figref{fig:truncation} in Appendix~\ref{app:capacity}), and the
\ZeroTruncationConfigs{} configurations that never truncate have $V$ effects
between \ZeroTruncationEffectMin{} and \ZeroTruncationEffectMax{}~pp. Truncation
is not the only route to a loss: DeepSeek V4.1 Flash has a $V$ effect of
\DeepSeekFlashVisibleEffect{}~pp at \DeepSeekFlashTruncation{}\% truncation.

Since every truncated episode fails, a correlation across configurations cannot
isolate the cap. A within-configuration sweep at sixteen times the cap does
(\figref{fig:capacity}b and Table~\ref{tab:capsweep} in Appendix~\ref{app:capsweep}). Run concurrently at
\SweepCapLow{} and \SweepCapHigh{} completion tokens with \SweepRepeats{}
repeats each, the \SweepBPassed{} truncation-heavy configurations, which
truncate \SweepBTruncLowMin{} to \SweepBTruncLowMax{}\% of their $V=128$
episodes at the low cap, truncate at most \SweepBTruncHighMax{}\% at the high
cap, and their pooled menu contrast shrinks from \SweepBPassedMenuLow{} to
\SweepBPassedMenuHigh{}~pp [\SweepBPassedMenuHighCILow{},
\SweepBPassedMenuHighCIHigh{}], an interaction of \SweepBPassedInteraction{}~pp
[\SweepBPassedInteractionCILow{}, \SweepBPassedInteractionCIHigh{}] that clears
the pre-specified 5~pp margin. The mild- and zero-truncation controls (strata M
and N) move by \SweepMInteraction{} and \SweepNInteraction{}~pp. The larger cap thus
removes most of their menu penalty, leaving \SweepBPassedMenuHigh{}~pp.

\begin{figure*}[t]
\centering
\includegraphics[width=\textwidth]{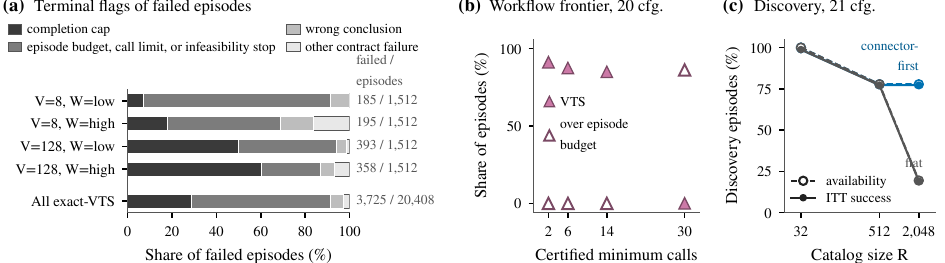}
\caption{Most failures end at a harness limit. (a) Terminal flags of failed episodes for the four $V\times W$ endpoint cells (raw and linked pooled) and for all \ExactVTSRows{} exactly graded episodes. The bars merge budget and infeasibility stops. (b) Frontier VTS (filled) and share of episodes over the \EpisodeTokenBudget{}-token episode budget (hollow) by certified calls, \FrontierModels{} configurations. (c) Discovery availability and ITT success by catalog size for two retrievers, means of configuration means.}
\label{fig:failures}
\Description{Panel a shows stacked horizontal bars of terminal flags (completion
cap, episode budget or call limit together with infeasibility stops, wrong conclusion,
other) for four
extreme cells and the whole population; harness-limit segments dominate every
bar. Panel b plots verified task success at four certified-call levels, flat near
ninety percent through fourteen calls and zero at thirty, and the share of
episodes over the token budget, zero through fourteen calls and most episodes
at thirty. Panel c plots availability and ITT success
against catalog size for two retrievers; the flat retriever collapses at the
largest catalog.}
\end{figure*}

\subsection{Where Do Failures End, and What Does the Record Reveal?}
\label{sec:q-end}

\paragraph{Failure census.}
Of the \ExactVTSRows{} exactly graded episodes, \TotalFailures{} fail
(\FailureRate{}\%, \figref{fig:failures}a): \CompletionCapFailures{} end at the completion cap (\CompletionCapShare{}\%),
\BudgetLimitFailures{} exceed the episode budget or call limit (\BudgetLimitShare{}\%),
\InfeasibilityFailures{} are stopped as infeasible (\InfeasibilityShare{}\%),
\WrongConclusionFailures{} reach a wrong conclusion (\WrongConclusionShare{}\%),
and \OtherContractFailures{} break another contract term
(\OtherContractShare{}\%). Harness limits thus end \HarnessLimitShare{}\% of
failures (\HarnessLimitShareNoInfeasibility{}\% without infeasibility stops). The model-run 30-call cells of the frontier and structural holdout contribute \ThirtyCallFailures{} failures,
nearly all over the episode budget, and in the four $V\times W$ endpoint cells
alone harness limits end \CapacityCellHarnessLimitShare{}\% of failures.
Obligation labels mislead on their own: all \StageSelection{} failures labeled
``selection'' come from episodes that made no call,
\FirstResponseTruncatedFailures{} of them cut off at the completion cap, so the
label reflects check order, not a wrong choice among tools.

\paragraph{Workflow frontier.}
When only the chain of handoffs grows, VTS over \FrontierModels{}
configurations is \FrontierVTSAtTwo{}\%, \FrontierVTSAtSix{}\%,
\FrontierVTSAtFourteen{}\%, and \FrontierVTSAtThirty{}\% at 2, 6, 14, and 30
certified calls (\figref{fig:failures}b). At 30 calls,
\FrontierThirtyBudgetExceeded{} of \FrontierEpisodesPerLevel{} episodes exceed
the episode budget and \FrontierThirtyInfeasible{} are stopped as infeasible,
and no episode executes more than \FrontierThirtyMaxCalls{} calls. The collapse
at 30 calls is where the \EpisodeTokenBudget{}-token episode budget runs out
(Appendix~\ref{app:frontier}).

\paragraph{Discovery.}
At $R=2{,}048$, the required tools are in only \AvailabilityFlatLarge{}\% of the
shortlists of the flat lexical retriever and in \AvailabilityRouterLarge{}\% of
those of the connector-first router (\figref{fig:failures}c), and
\DiscoveryUnavailableShare{}\% of all Discovery episodes fail before any model
call. The retriever does not depend on the model, so the catalog pressure acts
through retrieval (Appendix~\ref{app:discovery}).

\paragraph{Failure record.}
Does the failure record reveal the assigned pressure? On the
\LocalizationPackets{} configuration-family packets (a baseline cell and its three single-stress endpoint cells) in which the baseline succeeds and exactly one
stressed cell fails, a lookup decoder with one intent held out reaches
\BAJoint{}\% balanced accuracy from the failure record, above a constant guess
(\BAConstant{}\%) but below binned call counts alone (\BACallsOnly{}\%). Richer
encodings of the record reach \BATrace{}\% to \BACorrected{}\%, also below call
counts. Adding trace shape to the record gains \GainTraceVsCalls{}~pp
$[\GainTraceVsCallsLow, \GainTraceVsCallsHigh]$ over call counts (99\%
interval), and none of the five paired gains tested excludes zero
(Appendix~\ref{app:localization}).

\subsection{Which Repairs Help, and Do They Beat Switching?}
\label{sec:q-repair}

\begin{figure*}[t]
\centering
\includegraphics[width=\textwidth]{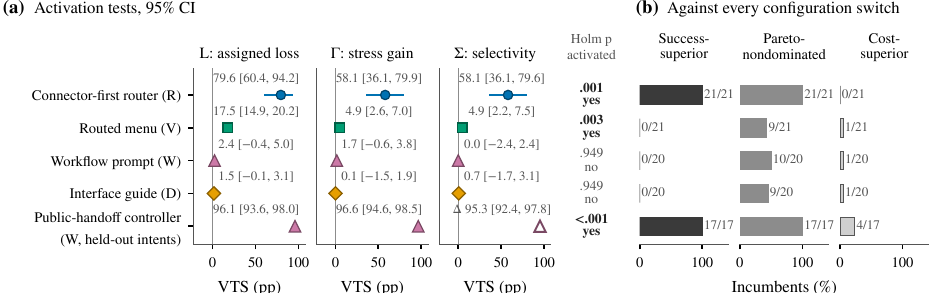}
\caption{Repairs that change which tools reach the model or who issues the calls
activate, prompt repairs do not, and only two repairs beat every
configuration switch on success. (a) $L$, $\Gamma$, and $\Sigma$ ($\Delta$ for the controller) in VTS pp (ITT success for the router) with 95\% cluster bootstrap intervals, and the activation decision. (b)
Share of incumbents for which each repair is success-superior,
Pareto-nondominated, or cost-superior against every configuration switch.}
\label{fig:repairs}
\Description{Panel a is a forest plot with five repair rows and columns for
assigned loss, stress gain, and selectivity (wrong-layer advantage for the
controller), plus an
activation gutter marking router, routed menu, and controller as activated. Panel
b shows, for each repair, bars for the share of incumbents that are
success-superior, Pareto-nondominated, and cost-superior, with counts printed.}
\end{figure*}

\paragraph{Activation.}
Three repairs activate (\figref{fig:repairs}a), and all three change which tools reach the model or who issues the calls. The connector-first router activates
(Holm $p$~\HolmEq{\RetrievalHolmP}) by raising availability (\figref{fig:failures}c). The routed menu
activates (Holm $p$~\HolmEq{\VisibleRepairHolmP}) with a stress gain of
$\Gamma=\VisibleRepairStressGain$~pp
$[\VisibleRepairStressGainCILow, \VisibleRepairStressGainCIHigh]$, and
three-run repeats on \SentinelFamilies{} families (sentinel repeats) give
$\Gamma=\SentinelRoutedGain$~pp $[\SentinelRoutedGainCILow, \SentinelRoutedGainCIHigh]$. The public-handoff controller raises VTS in
the structural holdout by \StructuralStressGain{}~pp
$[\StructuralStressGainCILow, \StructuralStressGainCIHigh]$ (Holm
$p$~\StructuralHolmP{}). It does so by taking execution away from the model. The
model issues only \ControllerThirtyModelCalls{} of the \ControllerThirtyCalls{}
executed calls, mean tokens per episode fall from \BaselineThirtyTokens{} to
\ControllerThirtyTokens{}, and episodes over budget fall from
\BaselineThirtyOverBudget{} to \ControllerThirtyOverBudget{} of
\BaselineThirtyRows{}. Its wrong-layer control, canonical delivery (records returned in a canonical encoding), has no detectable loss to repair
($L=\CanonicalControlAssignedLoss$~pp), so $\Delta$ is measured against an idle
control (Appendix~\ref{app:frontier}).

Neither prompt repair activates (Holm $p$~\HolmEq{\WorkflowPromptHolmP} for both), and both are tested on cells with no detectable loss (high $W$ and linked encoding).
Their stress gains are \WorkflowPromptStressGain{}~pp $[\WorkflowPromptStressGainCILow, \WorkflowPromptStressGainCIHigh]$ for the workflow prompt and \InterfaceGuideStressGain{}~pp $[\InterfaceGuideStressGainCILow, \InterfaceGuideStressGainCIHigh]$ for the interface guide, and two one-sided tests at the 5\% level \citep{schuirmann1987comparison} exclude stress gains beyond $\pm$\WorkflowPromptTOSTMargin{} and $\pm$\InterfaceGuideTOSTMargin{}~pp.

\paragraph{Comparison with switching.}
Against every configuration switch (\figref{fig:repairs}b), the routed menu is success-superior for \VisibleSuccessSuperior{} of \VisibleDecisionN{} incumbents and Pareto-nondominated for \VisiblePareto{}. 
Only two repairs beat every switch on success, and both do so without improving model behavior. 
The router is success-superior for all \RetrievalDecisionN{} incumbents because no configuration switch can fix retrieval. 
The controller is success-superior for all \StructuralModels{} incumbents and cost-superior for \StructuralCostSuperior{}, because it delegates execution. 
On capacity cells, a development pilot finds that a fixed switch beats every public-information repair (Appendix~\ref{app:pilot}).

\section{Discussion}
\label{sec:discussion}

\paragraph{Success scores should be read with their budgets.}
The visible menu is the only model-facing pressure with a large penalty, and
that penalty occurs mostly in configurations that run out of their
completion cap: a larger menu drives more episodes into the cap, almost always
before the first call. In the six truncation-heavy configurations, raising the cap
sixteenfold removes most of the menu penalty, from \SweepBPassedMenuLow{} to
\SweepBPassedMenuHigh{}~pp, and a small penalty remains. A success leaderboard under a large menu
ranks configurations partly by how they spend the completion cap.

\paragraph{Failure labels locate failures but do not explain them.}
A first-failure label such as ``tool selection'' reads as a diagnosis. Here most
failures end at a harness limit, every ``selection'' failure comes from an
episode that made no call, and our decoders recover the assigned pressure from
the record no better than from a count of calls. The label says where the
grader stopped, not why the agent failed. The limit that fired says how the
episode ended.

\paragraph{Recommendations.}
The repairs that activate change which tools reach the model or who issues the calls, yet only two, the router by construction and the controller by
delegation, beat every configuration switch on success. We recommend
reporting pressure effects with their harness budgets and each configuration's
truncation rate, reporting a first-failure label with the limit that fired, and
comparing a repair with switching the configuration on the same cells.
\section{Conclusion}
\label{sec:conclusion}

ToolBottleneckBench assigns one pressure at a time to fixed tasks under fixed
harness budgets and records the limit that ends each failed episode.
The visible menu is the only model-facing pressure with a large penalty. A sixteenfold
completion cap removes most of it in the six truncation-heavy configurations. Most failures end at a harness limit, and the
only repairs that beat every configuration switch do not improve model behavior. A tool agent's
success score measures its model and its harness budgets together.
Future work can vary the episode budget and call limit as we varied the completion cap, and test live tool servers.

\section*{Limitations}
\label{sec:limitations}

\paragraph{Budgets and repeats.}
The findings hold at the frozen harness budgets. The completion cap is the only
budget that varies; the sixteenfold sweep tests one high cap, \SweepCapHigh{}
tokens, on the routed-menu lane of \SweepConfigs{} configurations, so its
residual menu penalty is measured at the unchanged episode budget and call
limit (Appendix~\ref{app:crossover}). One nominal cap is not one effective
budget: providers count it differently, and reasoning effort stays at each provider's default on all routes but
one (Appendix~\ref{app:protocol}). Each core capacity
cell has one run per configuration; the family bootstrap absorbs run-to-run
variation, which the sentinel repeats (\SentinelDisagreement{}\% disagreement)
and the sweep repeats (\SweepDisagreementLow{}\% at \CompletionTokenCap{}
tokens) measure directly. The frontier spans \FrontierFamilies{} families.

\paragraph{Environment and release.}
The tasks are synthetic and graded exactly, without partial credit, and each
pressure effect includes everything its manipulation changes in practice
(Table~\ref{tab:pressures}). Prompts are fixed English texts. A configuration
couples a model with its provider route (one route is FP4-quantized;
Appendix~\ref{app:panel}), so its results describe that pair. The generator and its tasks are public
(Appendix~\ref{app:repro}), so models trained after the release may have seen
them.
\section*{Ethical Considerations}
All benchmark records, identities, and states are synthetic, so no personal
data reaches the model providers, and no human participants or annotators are
involved. Tools execute locally and have no external side effects. Results
describe configurations at fixed harness budgets and should not alone decide
procurement or deployment.
\section*{Use of Generative AI}
AI assistants were used for editing text, and for consistency checks between the manuscript and the released data, all under the authors' instruction. The authors designed the experiments, verified the claims and results, and take full responsibility for the content presented in the paper.
\ifdefined\BottleneckPreprint
\section*{Acknowledgments}
This work was supported by JST CREST Grant Number JPMJCR22M2 and JSPS KAKENHI Grant Numbers JP23K17456, JP23K28096, JP25H01117, JP25K21207, and JP26K03246.

\fi
\bibliography{references}

\clearpage

\appendix

\section{Harness Protocol and Prompts}
\label{app:protocol}

Table~\ref{tab:protocol} lists the frozen harness contract. Its budget rows hold for every block; its menu and prompt rows describe the capacity episodes and their prompt-repair arms. The numeric budgets are shared
by all configurations, but providers differ in which parameters they support and
in how they count tokens.

\begin{table}[t]
\centering\footnotesize
\caption{The frozen harness contract for capacity episodes and their matched
prompt-repair arms.}
\label{tab:protocol}
\setlength{\tabcolsep}{4pt}
\begin{tabular}{@{}>{\raggedright\arraybackslash}p{0.40\columnwidth}>{\raggedright\arraybackslash}p{0.52\columnwidth}@{}}
\toprule
Field & Setting \\
\midrule
Tool protocol & Native Chat Completions tools \\
Tool choice & Automatic; parallel-call flag false when sent \\
Temperature & Requested 0 unless the frozen profile omits it \\
Seed & Condition-bound unless the frozen profile omits it \\
Completion cap & \CompletionTokenCap{} tokens per request \\
Cap parameter & \texttt{max\_tokens} on Chat Completions routes; renamed \texttt{max\_output\_tokens} with the same value where a Responses transport was used; on routes that count reasoning tokens it bounds reasoning, visible text, and arguments together \\
Reasoning & Provider default on every route except one, which requested effort \texttt{high} (DeepSeek V4.1 Flash on DeepInfra); no thinking budget was set \\
Cap sweep & \SweepCapLow{} and \SweepCapHigh{} tokens per request through the same cap parameter, reasoning at the provider default on every route; \SweepConfigs{} configurations, \SweepRepeats{} repeats (Appendix~\ref{app:capsweep}) \\
Episode budget & \EpisodeTokenBudget{} reported tokens: prompt tokens, including cached ones, plus completion tokens \\
Call limit & Certified minimum $+\,2$ \\
Turn limit & Call limit $+\,2$ \\
Infeasibility stop & When success is infeasible under the call limit \\
Menu policy & Routed narrowing after the first executed operational call (oracle); paired fixed-menu lane \\
Prompt & Base English prompt; two declared repair variants \\
Provider fallback & Disabled; a served-model mismatch is an infrastructure failure \\
\bottomrule
\end{tabular}
\end{table}

\begin{figure*}[t]
\centering
\includegraphics[width=\textwidth]{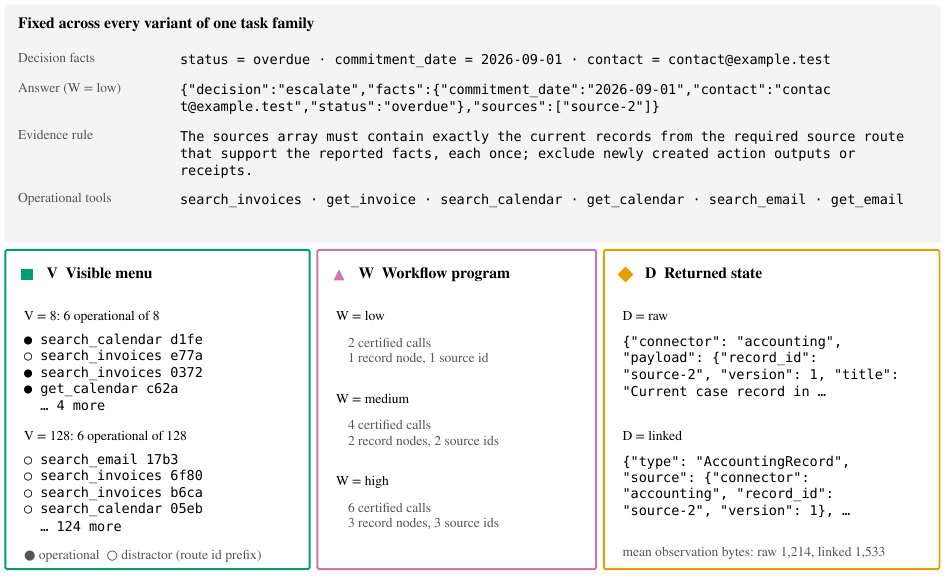}
\caption{One task family compiled under three pressures. The top block gives
the shared decision facts, reference answer, evidence rule, and six operational
tools. $V$ grows the menu from 8 to 128 schemas, six of them
operational. $W$ grows the certified program from one record node to three, and
the required \texttt{sources} array with it. $D$ re-encodes identical facts.
Identifiers and menus are abbreviated for space.}
\label{fig:variants}
\Description{A fixed goal block above three panels: the menu growing from 8 to
128 schemas, the certified program growing from one to three record nodes, and
the same record rendered as raw and as linked JSON.}
\end{figure*}

\paragraph{Example task family.}
\figref{fig:variants} shows one task family compiled under the $V$, $W$, and $D$
pressures.

\paragraph{System prompts.}
Capacity episodes and their prompt-repair arms use three variants of the system
prompt. Discovery, the frontier, the structural holdout, and ToolSandbox each
have their own frozen entry points and prompts. The base capacity prompt is the
same for the routed and fixed menus and reads:

\begin{quote}\small\ttfamily
You operate in a deterministic sandbox with synthetic business records. Use tools
iteratively and stop only when the task is supported by current evidence. Never
call an unlisted function. Each search/read description declares workspace,
region, and collection coordinates. Use only tools whose three coordinates exactly
match the required source route in the user request; other routes may contain
valid current records for other workspaces but cannot support this case. A
next\_handoff object identifies the connector and exact search query for the next
current record; follow it when more requested facts are still missing. Avoid
wasting the call budget on identical tool calls. Once all requested facts are
present, stop using tools and return raw JSON only, with no markdown fence or
explanatory prose. Never send messages, issue payment, or delete data. Follow the
user's exact JSON response schema.
\end{quote}

\noindent
The workflow prompt replaces the second sentence with ``Before every tool
request, privately verify the connector, operation, exact arguments, and
remaining public obligations; after the last tool, verify the evidence and
terminal state before answering.'' The interface guide keeps the second
sentence and inserts after it: ``Read returned records by interface: raw records
place values under payload; typed or linked records use type for the record
kind, place values under fields, and identify their origin under source.
Provenance maps JSON-pointer paths to source fields. Links and
relation\_dictionary summarize observed equality links but never replace reading
the cited source record.'' No other text differs between conditions.

\paragraph{Requests and sampling.}
The user message is the family's rendered request. Tools are supplied through
the provider's native tool interface with automatic tool choice. The client
sends \texttt{parallel\_tool\_calls=false}, the temperature, and the seed unless a
configuration's frozen execution profile omits them, and the released profiles
record each omission. If a model produces an invalid argument string
in an otherwise valid response, this counts as an agent failure, and the harness
never repairs it. Endpoint aliases and served versions are pinned.

\paragraph{Returned-state encodings.}
Five blocks of each observation vary across the encodings: \texttt{objects},
the field, type, and relation dictionaries, and \texttt{links}. The
\texttt{provenance} block and the source, hash, and tool blocks are identical
byte for byte across all four. Compact is larger than raw because it pays for a
field dictionary in every observation and saves a smaller amount on
\texttt{objects}.
\section{Obligations, Grading, and Audits}
\label{app:obligations}

The grader checks the nine obligations in the order given in
Table~\ref{tab:stages}, which also counts the first unmet obligation over the
exactly graded episodes. Answers are compared with Python value equality, where
\texttt{int} and \texttt{float} are compared after numeric coercion and
\texttt{str} is compared literally. Obligation $E$ requires the answer to cite
exactly the supporting record IDs. Every released record is at version one and
the answer carries no version, so $E$ compares record IDs. For tasks that
require an action, exactly one requested reversible change is permitted. For
tasks that do not, the state must stay unchanged while the mutating tool stays
visible. Governance $G$ fails when a tool other than the requested action changes state. Budget $B$ requires the episode to stay within the episode budget and the call limit. Completion $C$ fails when a response is cut off at the completion cap or breaks the response contract, or when the turn limit ends the episode.

\begin{table*}[t]
\centering
\footnotesize
\caption{First unmet obligation of each episode, in the order the grader
checks them, and the terminal flag of each failed episode. Population:
the \ExactVTSRows{} exactly graded episodes (\TotalFailures{} failures). All
\StageSelection{} ``selection'' labels belong to episodes without a call.}
\label{tab:stages}
\begin{tabular}{@{}lr@{\hspace{2em}}lrr@{}}
\toprule
First unmet obligation (emitted) & Episodes & Terminal flag (failures) & Count & Share (\%) \\
\midrule
availability & 0 & Completion cap & 1{,}068 & 28.7 \\
selection & 849 & Episode budget or call limit & 1{,}863 & 50.0 \\
workflow & 1{,}517 & Turn limit & 0 & 0.0 \\
answer & 1{,}340 & Infeasibility stop & 477 & 12.8 \\
evidence & 3 & Wrong conclusion & 204 & 5.5 \\
safety & 6 & Other contract failure & 113 & 3.0 \\
governance & 0 & All failures & 3{,}725 & 100.0 \\
budget & 0 &  &  &  \\
completion & 10 &  &  &  \\
success & 16{,}683 &  &  &  \\
\midrule
Scored episodes & 20{,}408 & & & \\
\bottomrule
\end{tabular}

\end{table*}

The analysis adds a termination-aware label: an episode truncated at the
completion cap before any call is labeled \texttt{completion\_no\_call} instead
of \texttt{selection}. The label is computed from call attempts and error codes
and leaves the grader's labels unchanged.

\paragraph{Audits.}
Static checks cover semantic signatures, the rank and support of the factorial
design, shortest programs, the layout of connectors and facts, nested menus,
task and action logic, and deterministic regeneration. Runtime checks decode
every encoding at every prefix of the reference program, validate dictionaries,
types, provenance, and links, exercise legal misses, accept every alternative
minimal call order, and reject altered availability, completion, budgets,
required steps, dependency order, trace hashes, trace evidence versions, answers,
sources, and final states. The public-handoff controller cannot see the hidden
program, the answer, or the grader. We test that its behavior does not depend on
the hidden program, that it follows the handoffs, and that it stops safely on an
invalid or empty handoff, over all \HoldoutIntents{} held-out 30-call tasks.

\paragraph{Zero-token rows.}
\ZeroTokenRows{} released rows report zero tokens. Of these,
\ZeroTokenAvailabilityRows{} stopped for lack of availability before any request,
and \ZeroTokenFailureRows{} are provider or execution failures. None has a
recorded provider response, so every row with a provider response reports its
tokens.
\section{Evaluation Panel}
\label{app:panel}

Table~\ref{tab:panel} lists the \CoreModels{} configurations, grouped by
organization, with the blocks each one ran, its baseline VTS, its
visible-menu and workflow effects, its truncation rate at $V=128$, and its cost.
All configurations use managed inference endpoints; one of them (DeepSeek V4
Flash 0731) is served through an FP4-quantized route.

\begin{table*}[t]
\centering
\footnotesize
\setlength{\tabcolsep}{3.5pt}
\caption{The \CoreModels{} configurations. Effects are per-configuration paired
within-family differences in VTS (pp) over \SemanticFamilies{} families, without
intervals; each routed $V$ effect averages 144 pairs. The routed $V$ effect is
the core contrast under the routed menu; the fixed $V$ effect is the
corresponding contrast on the fixed-menu lane (72 pairs per configuration, on the $(\mathrm{low},\mathrm{raw})$ and $(\mathrm{high},\mathrm{linked})$ cells),
with the full menu kept visible, i.e., minus the assigned loss $L$ of
the routed-menu repair. Truncation is the share of $V=128$ capacity episodes
that end at the completion cap. Cost is the token cost at frozen route prices. Blocks (Table~\ref{tab:populations}): C core
capacity and Discovery; A prompt repairs, sentinel repeats, and frontier; S structural
holdout ($^{*}$ confirmatory plan); M concurrent ToolSandbox menu; E historical
ToolSandbox rank and menu. Baseline VTS pools the $(8,\mathrm{low},\mathrm{raw})$ and $(8,\mathrm{low},\mathrm{linked})$ cells.}
\label{tab:panel}
\resizebox{\textwidth}{!}{\setlength{\tabcolsep}{3pt}
\begin{tabular}{@{}llcrrrrrr@{}}
\toprule
Configuration & Org. & Blocks & \begin{tabular}[b]{@{}r@{}}Base\\ VTS\end{tabular} & \begin{tabular}[b]{@{}r@{}}$V$ eff.\\ routed\end{tabular} & \begin{tabular}[b]{@{}r@{}}$V$ eff.\\ fixed\end{tabular} & $W$ eff. & Trunc. & USD \\
\midrule
Claude Haiku 5.5 & Anthropic & CASM & 94.4 & \ensuremath{-}7.6 & \ensuremath{-}2.8 & \ensuremath{-}6.2 & 6.2 & 3.44 \\
Claude Opus 5 & Anthropic & CASM & 94.4 & 1.4 & \ensuremath{-}1.4 & 1.4 & 0.0 & 165.18 \\
Claude Opus 5.5 & Anthropic & CASM & 97.2 & 0.0 & 1.4 & 0.0 & 0.0 & 130.52 \\
Claude Sonnet 5 & Anthropic & CASM & 93.1 & \ensuremath{-}9.0 & \ensuremath{-}4.2 & \ensuremath{-}11.8 & 9.0 & 66.45 \\
Claude Sonnet 5.5 & Anthropic & CASM & 94.4 & 0.0 & \ensuremath{-}1.4 & 1.4 & 0.0 & 66.51 \\
\addlinespace[1.5pt]
GPT-5.6 Luna & OpenAI & CAE & 95.8 & 2.8 & \ensuremath{-}16.7 & \ensuremath{-}2.8 & 0.0 & 1.81 \\
GPT-5.6 Sol & OpenAI & CASM & 100.0 & \ensuremath{-}2.1 & \ensuremath{-}1.4 & \ensuremath{-}2.1 & 0.0 & 84.33 \\
GPT-5.6 Terra & OpenAI & CAE & 97.2 & \ensuremath{-}2.1 & \ensuremath{-}11.1 & 0.7 & 0.0 & 8.15 \\
GPT-6 Luna & OpenAI & CASM & 93.1 & 0.0 & \ensuremath{-}4.2 & 0.0 & 0.0 & 2.13 \\
GPT-6.1 Sol & OpenAI & CASM & 100.0 & 0.0 & 0.0 & 0.0 & 0.0 & 40.22 \\
\addlinespace[1.5pt]
Gemini 3.6 Flash & Google & CAE & 93.1 & 5.6 & \ensuremath{-}18.1 & \ensuremath{-}30.6 & 6.2 & 20.02 \\
Gemini 3.8 Flash & Google & CASM & 87.5 & \ensuremath{-}52.1 & \ensuremath{-}50.0 & \ensuremath{-}6.2 & 67.4 & 25.01 \\
\addlinespace[1.5pt]
Qwen3.7 Flash & Alibaba & CAS$^{*}$E & 88.9 & \ensuremath{-}34.7 & \ensuremath{-}58.3 & \ensuremath{-}6.9 & 45.8 & 0.88 \\
Qwen3.8 Flash & Alibaba & C & 66.7 & \ensuremath{-}16.7 & \ensuremath{-}34.7 & 5.6 & 27.8 & 1.92 \\
Qwen3.8 Max (0902) & Alibaba & CASM & 73.6 & \ensuremath{-}4.9 & 8.3 & 18.8 & 5.6 & 58.14 \\
\addlinespace[1.5pt]
DeepSeek V4 Flash 0731 & DeepSeek & CAS$^{*}$E & 93.1 & \ensuremath{-}11.1 & \ensuremath{-}15.3 & 2.8 & 4.2 & 2.48 \\
DeepSeek V4 Pro 0813 & DeepSeek & CASM & 94.4 & \ensuremath{-}24.3 & \ensuremath{-}41.7 & \ensuremath{-}10.4 & 27.8 & 33.30 \\
DeepSeek V4.1 Flash & DeepSeek & CASM & 69.4 & \ensuremath{-}18.1 & \ensuremath{-}20.8 & 23.6 & 0.7 & 4.26 \\
\addlinespace[1.5pt]
HY4 Preview & Tencent & CAS & 27.8 & \ensuremath{-}2.8 & \ensuremath{-}6.9 & 34.7 & 0.7 & 19.83 \\
\addlinespace[1.5pt]
GLM-5.3 & Z.ai & CAS & 93.1 & \ensuremath{-}53.5 & \ensuremath{-}59.7 & 6.2 & 56.2 & 31.99 \\
GLM-5.3 Flash & Z.ai & CAS & 95.8 & \ensuremath{-}28.5 & \ensuremath{-}29.2 & \ensuremath{-}0.7 & 27.8 & 1.81 \\
\bottomrule
\end{tabular}
}
\end{table*}

Each analysis uses its own population (Table~\ref{tab:populations}). The
blocks differ in configurations, resampling unit, and success measure: exact
VTS for the capacity, repair, sentinel, frontier, and structural blocks;
ITT success for Discovery; and ToolSandbox's own evaluator for
the external blocks.

\begin{table}[t]
\centering
\footnotesize
\setlength{\tabcolsep}{3pt}
\caption{Analysis populations: configurations, rows, and success measure
(VTS, ITT success $A\cdot S$, or ToolSandbox's evaluator) of each block.}
\label{tab:populations}
\setlength{\tabcolsep}{3pt}
\begin{tabular}{@{}lrrl@{}}
\toprule
Block & Cfg. & Rows & Measure \\
\midrule
Capacity factorial & 21 & 12{,}096 & VTS \\
Discovery ($R$) & 21 & 4{,}536 & ITT \\
Prompt repairs & 20 & 2{,}880 & VTS \\
Sentinel repeats & 20 & 1{,}440 & VTS \\
Workflow frontier & 20 & 320 & VTS \\
Structural holdout & 17 & 3{,}672 & VTS \\
TS menu, concurrent & 12 & 1{,}152 & evaluator \\
TS menu, historical & 5 & 320 & evaluator \\
TS rank & 5 & 160 & evaluator \\
\midrule
All blocks & 21 & 26{,}576 & \\
Exact-VTS population & & 20{,}408 & VTS \\
\bottomrule
\end{tabular}

\end{table}
\section{Statistical Analysis and Capacity Sensitivity}
\label{app:capacity}

Table~\ref{tab:capacity} gives every capacity contrast under the three
resampling units. The results in the main text use families as the primary unit.

\begin{table*}[t]
\centering
\footnotesize
\setlength{\tabcolsep}{4pt}
\caption{Capacity contrasts under three resampling units. Estimates are paired
within-family differences in VTS (pp) over \CoreModels{} configurations, averaging
\ContrastPairs{} family-by-configuration differences per contrast. Intervals are 95\% cluster bootstraps with
\BootstrapDraws{} draws over \SemanticFamilies{} families, \Intents{} intents, and
\CoreModels{} configurations, each with its own Holm correction over the eight
contrasts. The last column is the exact sign-flip test over intents. Intervals are
unadjusted percentile intervals and $p$ values come from the centered
bootstrap, so the two can disagree near zero. $<.001$
marks the resolution floor of the bootstrap.}
\label{tab:capacity}
\setlength{\tabcolsep}{3pt}
\begin{tabular}{@{}lr cc cc ccc@{}}
\toprule
 & & \multicolumn{2}{c}{36 families} & \multicolumn{2}{c}{21 configurations} & \multicolumn{3}{c}{12 intents} \\
\cmidrule(lr){3-4}\cmidrule(lr){5-6}\cmidrule(l){7-9}
Contrast & Est. & 95\% CI & Holm $p$ & 95\% CI & Holm $p$ & 95\% CI & Holm $p$ & exact Holm $p$ \\
\midrule
$V$: 128 $-$ 8 & \ensuremath{-}12.3 & [\ensuremath{-}14.6, \ensuremath{-}9.9] & \ensuremath{<}.001 & [\ensuremath{-}19.8, \ensuremath{-}5.7] & .011 & [\ensuremath{-}14.7, \ensuremath{-}9.7] & \ensuremath{<}.001 & .004 \\
$W$: high $-$ low & 0.8 & [\ensuremath{-}1.2, 2.8] & .836 & [\ensuremath{-}4.6, 6.5] & 1.000 & [\ensuremath{-}1.2, 3.1] & .624 & .949 \\
$D$: compact $-$ raw & \ensuremath{-}3.7 & [\ensuremath{-}5.7, \ensuremath{-}1.9] & .003 & [\ensuremath{-}5.0, \ensuremath{-}2.5] & \ensuremath{<}.001 & [\ensuremath{-}5.7, \ensuremath{-}1.7] & .002 & .068 \\
$D$: typed $-$ raw & \ensuremath{-}0.7 & [\ensuremath{-}2.4, 0.9] & .836 & [\ensuremath{-}2.6, 1.1] & 1.000 & [\ensuremath{-}1.7, 0.1] & .624 & .949 \\
$D$: linked $-$ raw & \ensuremath{-}1.4 & [\ensuremath{-}2.9, 0.1] & .301 & [\ensuremath{-}2.8, 0.0] & .295 & [\ensuremath{-}2.7, 0.03] & .177 & .386 \\
$V{\times}W$ & 3.0 & [0.2, 5.7] & .202 & [\ensuremath{-}0.1, 6.1] & .342 & [0.5, 5.2] & .070 & .252 \\
$V{\times}$linked & \ensuremath{-}1.3 & [\ensuremath{-}3.5, 1.0] & .780 & [\ensuremath{-}3.5, 1.2] & 1.000 & [\ensuremath{-}3.4, 1.0] & .624 & .949 \\
$W{\times}$linked & 1.7 & [\ensuremath{-}0.8, 4.1] & .762 & [\ensuremath{-}1.3, 4.8] & 1.000 & [0.1, 3.1] & .177 & .386 \\
\bottomrule
\end{tabular}

\end{table*}

\begin{figure*}[t]
\centering
\includegraphics[width=\textwidth]{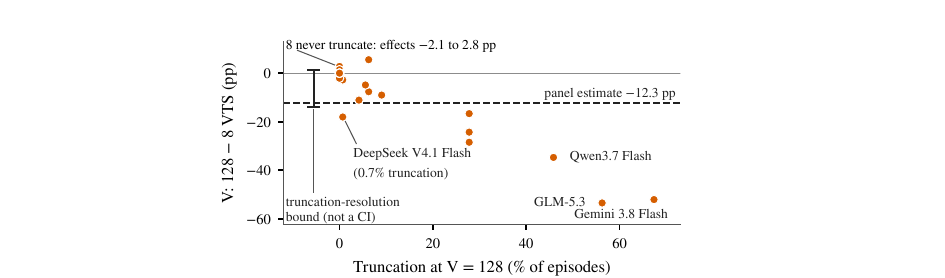}
\caption{The visible-menu penalty moves with completion-cap truncation. Each
configuration's $V$ effect against its truncation rate at $V=128$, with the panel
estimate (dashed) and the truncation-resolution bound (bracket, not a confidence
interval).}
\label{fig:truncation}
\Description{A scatter of the per-configuration visible-menu effect against the
truncation rate at 128 schemas, falling from near zero at no truncation to about
minus fifty points at the highest truncation rates, with named points for
GLM-5.3, Gemini 3.8 Flash, Qwen3.7 Flash, DeepSeek V4.1 Flash, and the
zero-truncation cluster.}
\end{figure*}

\paragraph{Design cells and contrast weights.}
The eight endpoint cells are
$\mathcal C=\{8,128\}\times\{\mathrm{low},\mathrm{high}\}\times
\{\mathrm{raw},\mathrm{linked}\}$, and the four checkpoint cells are
$(32,\mathrm{low},\mathrm{raw})$, $(8,\mathrm{medium},\mathrm{raw})$,
$(8,\mathrm{low},\mathrm{compact})$, and $(8,\mathrm{low},\mathrm{typed})$. For an
endpoint cell $c$ with mean success $m_c$, let $s_a(c)$ be $-1$ at the low level
and $+1$ at the high level of axis $a$. Then
\begin{align*}
E_a&=\tfrac14\sum_{c\in\mathcal C}s_a(c)m_c,\\
E_{ab}&=\tfrac12\sum_{c\in\mathcal C}s_a(c)s_b(c)m_c,
\end{align*}
so each main effect averages four paired differences and each interaction two
differences-in-differences. Compact and typed have weight $+1$ at their
checkpoint and $-1$ at $(8,\mathrm{low},\mathrm{raw})$. Curvature at the
midpoints and the three-way interaction are exploratory and not part of the
eight tests.

\paragraph{Resampling units.}
With configurations as the resampling unit, the visible menu is
$[\VisibleConfigCILow, \VisibleConfigCIHigh]$ (Holm
$p$~\HolmEq{\VisibleConfigHolmP}), workflow pressure
$[\WorkflowConfigCILow, \WorkflowConfigCIHigh]$ (Holm
$p$~\HolmEq{\WorkflowConfigHolmP}), and compact encoding
$[\CompactConfigCILow, \CompactConfigCIHigh]$ (Holm $p$~\CompactConfigHolmP{}).
A centered percentile bootstrap with \BootstrapDraws{} draws cannot resolve a
two-sided $p$ below $10^{-4}$, which the Holm correction over eight contrasts
raises to \BootstrapPFloor{}, so every entry printed as $<.001$ sits at that
floor. With \Intents{} intent clusters, we also report the
exact sign-flip test. For compact encoding it gives $p=\CompactExactIntentP$
before correction and Holm $p$~\HolmEq{\CompactExactIntentHolmP}.

\begin{figure*}[t]
\centering
\includegraphics[width=\textwidth]{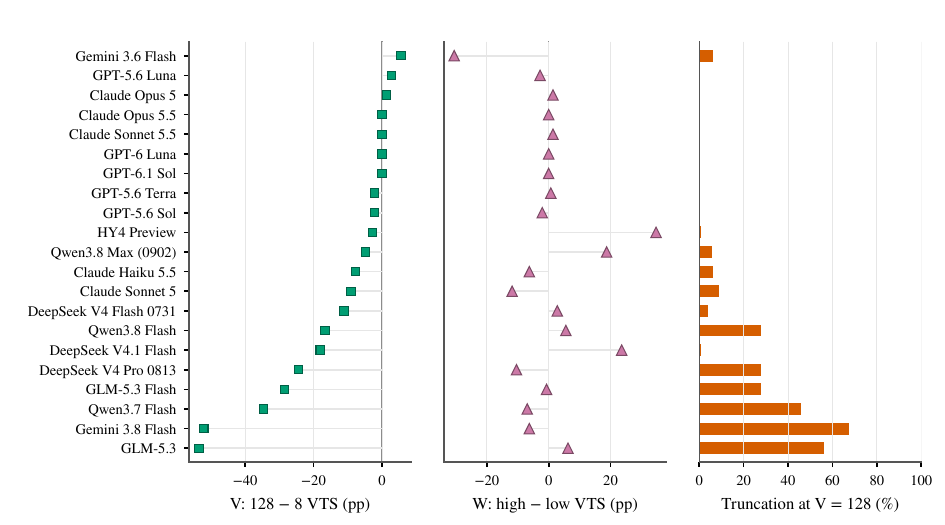}
\caption{Per-configuration effects differ in size and sign, and large $V$
penalties occur with high truncation. Within-configuration paired differences
in VTS (pp) for $V$ (128 minus 8, routed menu) and $W$ (high minus low), with the $V=128$
truncation rate in a side column, for \CoreModels{} configurations over
\SemanticFamilies{} families, sorted by the $V$ effect. Points carry no
interval.}
\label{fig:config-effects}
\Description{Two forest-style panels with one marker per configuration for the
visible-menu and workflow effects, sorted by the visible effect, beside a column
of truncation rates.}
\end{figure*}

\paragraph{Heterogeneity across configurations.}
Some configurations have zero variance across families (GPT-6.1 Sol for every
contrast), so a Der\-Simonian--Laird synthesis
\citep{dersimonian1986metaanalysis} is undefined; \figref{fig:config-effects} instead shows the per-configuration effects of $V$ and $W$ with each configuration's
truncation rate. Reweighting so that each organization counts equally gives
\OrgBalancedVisibleEffect{}~pp for the visible menu,
\OrgBalancedWorkflowEffect{}~pp for workflow pressure, and
\OrgBalancedCompactEffect{}~pp for compact encoding.

\paragraph{Interaction.}
If the two pressures acted independently, the expected $V\times W$ interaction
would be \InteractionIndependenceBenchmark{}~pp. Of the observed coefficient, \InteractionFloorPatternShare{}\% comes from \InteractionFloorUnits{} of
\InteractionUnits{} units in which every stressed cell fails. Dropping the units
whose sign is forced by the single-stress cells gives
\InteractionNonFloorEffect{}~pp
$[\InteractionNonFloorCILow, \InteractionNonFloorCIHigh]$. Joint stress succeeds while a single stress fails in \NonMonotoneUnits{} units (\NonMonotoneShare{}\%).

\paragraph{Truncation bounds.}
The truncation-resolution bound holds every non-truncated episode and its
outcome fixed under the frozen cap; the sweep (Appendix~\ref{app:capsweep})
measures what a larger cap does. The two ends of the bound assume that every truncated failure at
$V=8$ would have succeeded and every one at $V=128$ failed, and the reverse,
with each arm resolving independently. Adding the assumption that both arms
resolve alike gives $[\CommonResolutionLower, \CommonResolutionUpper]$~pp. For
comparison, the two workflow arms differ by only \WorkflowArmTruncationGap{}
truncated episodes, while the two menu arms have \TruncatedFailuresVEight{}
and \TruncatedFailuresVOneTwoEight{} truncated failures.

\paragraph{Infrastructure failures and the routed-menu lane.}
\InfrastructureConditions{} conditions failed for infrastructure reasons
(\InfrastructureCapacityConditions{} capacity and
\InfrastructureDiscoveryConditions{} Discovery). They are kept and counted as
failures of their configuration. Over \RoutingPairs{} paired episodes, the
routed-menu lane exceeds the fixed-menu lane by \RoutingGain{}~pp
$[\RoutingGainCILow, \RoutingGainCIHigh]$, and the fixed-menu lane uses
\RoutingTokenSaving{} more tokens and \RoutingSchemaExposureSaving{} more schema
exposures per episode.
\section{Completion-Cap Crossover and Sweep}
\label{app:crossover}

Two studies vary the completion cap per request while the episode budget and
the call limit stay fixed: a crossover at four times the cap and a sweep at
sixteen times.

\paragraph{Crossover.}
The crossover uses three strata defined by the panel's $V=128$ truncation share and menu penalty: truncation-heavy (B, truncation at least
20\%), loss without much truncation (S, truncation at most 15\% with a loss of
at least 8~pp), and null (N). The \CrossoverConfigs{} configurations are GLM-5.3
Flash and Qwen3.8 Flash (B), DeepSeek V4 Flash 0731 (S, on a different provider
route from its panel counterpart, so both of its arms come from the crossover),
and GPT-5.6 Luna (N). Each ran the 432-condition $V\times W\times D$ capacity
factorial on the routed-menu lane at \CompletionTokenCap{} and
\CrossoverCapHigh{} completion tokens, twice, with the arm order
counterbalanced across repeats. The estimand is the paired within-family menu
contrast in each arm and the difference between the arms, per stratum. Of
\CrossoverRowsRead{} rows read, \CrossoverRowsKept{} were kept; the other
\CrossoverRowsExcluded{} had a provider failure or missing usage data.

\begin{figure*}[t]
\centering
\includegraphics[width=\textwidth]{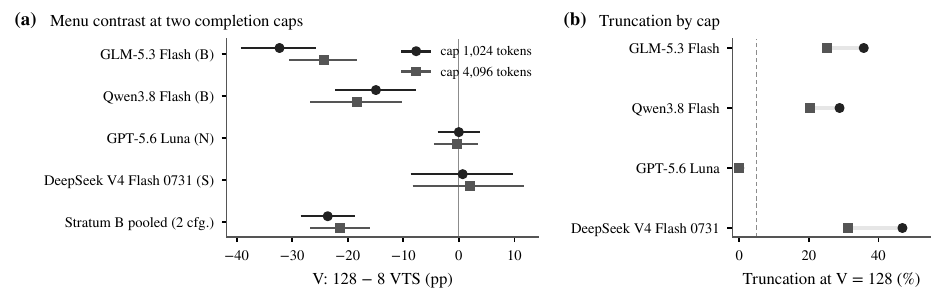}
\caption{The menu penalty of the truncation-heavy configurations persists at four
times the cap. (a) Visible-menu contrast (pp) at \CompletionTokenCap{} and
\CrossoverCapHigh{} completion tokens for each of the \CrossoverConfigs{}
configurations and pooled over stratum B, with 95\% family-cluster bootstrap
intervals. (b) Share of $V=128$ episodes truncated at each cap; the dashed line
is the \CrossoverManipulationCeiling{}\% ceiling of the manipulation check.}
\label{fig:crossover}
\Description{Two panels. (a) Paired points per configuration and for the pooled
stratum B at the two caps; the two truncation-heavy configurations stay far
below zero at both caps, while the null configuration stays near zero. (b)
Truncation shares at the two caps per configuration, all but the null
configuration far above the dashed 5 percent line.}
\end{figure*}

At \CrossoverCapHigh{} tokens both stratum-B configurations still truncate
\CrossoverTruncBCapHighMin{} to \CrossoverTruncBCapHighMax{}\% of their $V=128$
episodes, down from \CrossoverTruncBCapLowMin{} to
\CrossoverTruncBCapLowMax{}\%, above the \CrossoverManipulationCeiling{}\%
manipulation-check ceiling, so Table~\ref{tab:crossover} and
\figref{fig:crossover} describe the penalty under persisting truncation. The
stratum-B pooled menu contrast is \CrossoverBMenuCapLow{}~pp
$[\CrossoverBMenuCapLowCILow, \CrossoverBMenuCapLowCIHigh]$ at the lower cap and
\CrossoverBMenuCapHigh{}~pp
$[\CrossoverBMenuCapHighCILow, \CrossoverBMenuCapHighCIHigh]$ at the higher cap,
an interaction of \CrossoverBInteraction{}~pp
$[\CrossoverBInteractionFamilyCILow, \CrossoverBInteractionFamilyCIHigh]$;
letting each excluded row take either outcome keeps the interaction within
$[\CrossoverBInteractionBoundLow, \CrossoverBInteractionBoundHigh]$~pp. The null
stratum moves by \CrossoverNInteraction{}~pp
$[\CrossoverNInteractionCILow, \CrossoverNInteractionCIHigh]$ and the stratum-S
configuration by \CrossoverSInteraction{}~pp
$[\CrossoverSInteractionCILow, \CrossoverSInteractionCIHigh]$; the latter
truncates heavily at both caps yet shows no menu penalty. The two repeats disagree on
\CrossoverDisagreementCapLow{}\% of conditions at \CompletionTokenCap{} tokens
and \CrossoverDisagreementCapHigh{}\% at \CrossoverCapHigh{}.
\begin{table*}[t]
\centering
\footnotesize
\setlength{\tabcolsep}{3.5pt}
\caption{A fourfold completion cap left the truncation-heavy configurations
truncating, and their menu penalty changed little. Paired within-family
visible-menu contrast in VTS (pp) at each cap and the difference, per
configuration and stratum, with $V=128$ truncation rates. \CrossoverConfigs{}
configurations, \CrossoverRepeats{} repeats per cap, \CrossoverRowsKept{} kept
rows, 95\% family-cluster bootstrap intervals; Dis.\ is the share of conditions
whose outcome differs between the two repeats, both caps pooled. Superscripts
give the stratum (B, S, N). Truncation persists at the higher cap, so no row
estimates a menu effect free of truncation.}
\label{tab:crossover}
\resizebox{\textwidth}{!}{\setlength{\tabcolsep}{3pt}
\begin{tabular}{@{}lrrlllr@{}}
\toprule
 & \multicolumn{2}{c}{Trunc.\ (\%)} & \multicolumn{2}{c}{Menu contrast (pp)} & & \multicolumn{1}{c}{Dis.} \\
\cmidrule(lr){2-3}\cmidrule(lr){4-5}
Config.$^{\mathrm{stratum}}$ & 1{,}024 & 4{,}096 & at 1{,}024 & at 4{,}096 & Interaction & (\%) \\
\midrule
GLM-5.3 Flash$^{B}$ & 35.8 & 25.3 & \ensuremath{-}32.3\,[\ensuremath{-}39.2,\,\ensuremath{-}25.7] & \ensuremath{-}24.3\,[\ensuremath{-}30.6,\,\ensuremath{-}18.4] & 8.0\,[2.1,\,14.2] & 19.4 \\
Qwen3.8 Flash$^{B}$ & 28.8 & 20.2 & \ensuremath{-}14.9\,[\ensuremath{-}22.2,\,\ensuremath{-}7.6] & \ensuremath{-}18.3\,[\ensuremath{-}26.8,\,\ensuremath{-}10.3] & \ensuremath{-}3.9\,[\ensuremath{-}11.6,\,3.8] & 24.8 \\
GPT-5.6 Luna$^{N}$ & 0.0 & 0.0 & 0.0\,[\ensuremath{-}3.8,\,3.8] & \ensuremath{-}0.3\,[\ensuremath{-}4.5,\,3.5] & \ensuremath{-}0.3\,[\ensuremath{-}4.5,\,3.8] & 6.9 \\
DeepSeek V4 Flash 0731$^{S}$ & 46.9 & 31.2 & 0.7\,[\ensuremath{-}8.7,\,9.7] & 2.1\,[\ensuremath{-}8.3,\,11.8] & 1.4\,[\ensuremath{-}8.3,\,11.1] & 39.8 \\
\midrule
Stratum B (2 cfg.) & & & \ensuremath{-}23.6\,[\ensuremath{-}28.5,\,\ensuremath{-}18.8] & \ensuremath{-}21.3\,[\ensuremath{-}26.7,\,\ensuremath{-}16.1] & 2.1\,[\ensuremath{-}2.6,\,6.8] & \\
Stratum S (1 cfg.) & & & 0.7\,[\ensuremath{-}8.7,\,9.7] & 2.1\,[\ensuremath{-}8.3,\,11.8] & 1.4\,[\ensuremath{-}8.3,\,11.1] & \\
Stratum N (1 cfg.) & & & 0.0\,[\ensuremath{-}3.8,\,3.8] & \ensuremath{-}0.3\,[\ensuremath{-}4.5,\,3.5] & \ensuremath{-}0.3\,[\ensuremath{-}4.5,\,3.8] & \\
All (4 cfg.) & & & \ensuremath{-}11.6\,[\ensuremath{-}14.7,\,\ensuremath{-}8.8] & \ensuremath{-}10.2\,[\ensuremath{-}13.7,\,\ensuremath{-}7.0] & 1.3\,[\ensuremath{-}1.5,\,4.1] & \\
\bottomrule
\end{tabular}
}
\end{table*}

\paragraph{What fills a truncated first response.}
For the
\TruncatedFirstConfigs{} configurations whose provider usage payload
reports reasoning tokens, the saved provider responses show the first response of every episode that carried 128 schemas
(about \TruncatedFirstRequestsPerConfig{} such requests per configuration, over
all blocks; Table~\ref{tab:truncated-first}). Of the \TruncatedFirstTotal{} first
responses that ended at the cap, \TruncatedFirstReasoningOnly{}
(\TruncatedFirstReasoningOnlyShare{}\%) contain no visible text and no tool
call: their completion tokens are reasoning tokens, so the cap was spent before
any action. Most of them come from one configuration, Gemini 3.8 Flash. Of the rest, \TruncatedFirstPartialCall{} end inside a tool call and \TruncatedFirstPartialText{} inside visible text. Among the \TruncatedFirstConfigs{} configurations, \TruncatedFirstZeroConfigs{} never truncate a first response.

\begin{table*}[t]
\centering
\footnotesize
\caption{Truncated first responses to 128-schema requests are mostly reasoning
tokens only. First requests, those ending at the completion cap, and the content
of the truncated response, for the \TruncatedFirstConfigs{} configurations with
reasoning-token telemetry; counts from the saved provider responses over all blocks.}
\label{tab:truncated-first}
\setlength{\tabcolsep}{3pt}
\begin{tabular}{@{}lrrrrrr@{}}
\toprule
 & & & \multicolumn{4}{c}{Content of the truncated response} \\
\cmidrule(l){4-7}
Configuration & First req. & Truncated & Reasoning only & Partial call & Partial text & Both \\
\midrule
Claude Haiku 5.5 & 240 & 15 & 11 & 4 & 0 & 0 \\
Claude Opus 5 & 241 & 0 & 0 & 0 & 0 & 0 \\
Claude Opus 5.5 & 240 & 0 & 0 & 0 & 0 & 0 \\
Claude Sonnet 5.5 & 240 & 0 & 0 & 0 & 0 & 0 \\
GPT-6 Luna & 240 & 0 & 0 & 0 & 0 & 0 \\
GPT-6.1 Sol & 240 & 0 & 0 & 0 & 0 & 0 \\
Gemini 3.8 Flash & 240 & 142 & 95 & 35 & 12 & 0 \\
Qwen3.8 Max (0902) & 240 & 13 & 13 & 0 & 0 & 0 \\
DeepSeek V4.1 Flash & 240 & 1 & 1 & 0 & 0 & 0 \\
\bottomrule
\end{tabular}

\end{table*}

\subsection{Sixteenfold Completion-Cap Sweep}
\label{app:capsweep}

The sweep repeats the design at sixteen times the cap, \SweepCapHigh{}
completion tokens per request against the frozen \SweepCapLow{}, on \SweepConfigs{} panel configurations. The high cap is the largest power of two that keeps the unchanged \EpisodeTokenBudget{}-token episode budget from binding on the episodes the sweep is meant to rescue. Strata, defined separately from those of the crossover, follow the panel's $V=128$ truncation share: B, truncation-heavy, at least 20\% (Gemini
3.8 Flash, GLM-5.3, Qwen3.7 Flash, Qwen3.8 Flash, GLM-5.3 Flash, and DeepSeek
V4 Pro 0813, every panel configuration that meets the rule); M, 5 to 20\% (Claude Haiku 5.5); and N, no truncation and a
$V$ effect within 3~pp of zero (GPT-6 Luna). Every configuration kept the route and request parameters of its panel runs.

Each configuration ran the 432-condition capacity factorial on the
routed-menu lane at both caps, the arms concurrent, \SweepRepeats{} repeats per
arm; all \SweepRowsRead{} rows read were kept. The primary estimand is the
paired within-family visible-menu contrast ($V=128$ minus $V=8$) in each arm
over the four endpoint $W\times D$ cells and its difference between the arms,
pooled over the stratum-B configurations that pass the manipulation check;
strata M and N are controls.

\paragraph{Manipulation check.}
In the \SweepCapLow{}-token arm the stratum-B configurations truncated
\SweepBTruncLowMin{} to \SweepBTruncLowMax{}\% of their $V=128$ episodes; at
\SweepCapHigh{} tokens the share was at most \SweepBTruncHighMax{}\%
(Table~\ref{tab:capsweep}), under the 5\% ceiling for every one, so all
\SweepBPassed{} enter the primary reading. The controls truncated
\SweepClaudeHaikuFiveFiveTruncLow{}\% (M) and \SweepGPTSixLunaTruncLow{}\% (N)
at the low cap and none at the high cap. The low-cap contrast of each stratum-B
configuration differs from its released $V$ effect by \SweepBDriftMin{} to
\SweepBDriftMax{}~pp, a larger penalty in every case (Rel.\ column of
Table~\ref{tab:capsweep}). For all but one stratum-B configuration, the larger penalty comes with a higher truncation share at the same cap (Trunc.\ columns of Tables~\ref{tab:capsweep} and~\ref{tab:panel}); all \SweepBPassed{} low-cap points fall within the spread of the panel's effect--truncation relation (\figref{fig:truncation}). The primary reading uses only the within-sweep interaction.

\paragraph{Outcome.}
Pooled over the \SweepBPassed{} truncation-heavy configurations, the paired
visible-menu contrast is \SweepBPassedMenuLow{}~pp
[\SweepBPassedMenuLowCILow{}, \SweepBPassedMenuLowCIHigh{}] at \SweepCapLow{}
tokens and \SweepBPassedMenuHigh{}~pp [\SweepBPassedMenuHighCILow{},
\SweepBPassedMenuHighCIHigh{}] at \SweepCapHigh{}, an interaction of
\SweepBPassedInteraction{}~pp [\SweepBPassedInteractionCILow{},
\SweepBPassedInteractionCIHigh{}]. The interaction interval lies entirely above
the pre-specified 5~pp margin and the high-cap interval excludes zero. In these
configurations, the completion cap accounts for most of the menu penalty, and a
small penalty remains. Each
stratum-B configuration's own interaction interval also lies above the margin
(Table~\ref{tab:capsweep}).
The controls move within noise: M from \SweepMMenuLow{} to
\SweepMMenuHigh{}~pp (interaction \SweepMInteraction{}
[\SweepMInteractionCILow{}, \SweepMInteractionCIHigh{}]) and N from
\SweepNMenuLow{} to \SweepNMenuHigh{}~pp (\SweepNInteraction{}
[\SweepNInteractionCILow{}, \SweepNInteractionCIHigh{}]). The two repeats
disagree on \SweepDisagreementLow{}\% of conditions at the low cap and
\SweepDisagreementHigh{}\% at the high cap.

\begin{table*}[t]
\centering
\footnotesize
\setlength{\tabcolsep}{3.5pt}
\caption{A sixteenfold completion cap cuts the truncation to at most
\SweepBTruncHighMax{}\% and takes away most of the menu penalty. Share of $V=128$ episodes truncated at each cap; paired
within-family visible-menu contrast in VTS (pp) over the four endpoint
$W\times D$ cells at each cap, with 95\% family-cluster bootstrap intervals, and their
difference (interaction); the same configuration's released $V$ effect (Rel.,
Table~\ref{tab:panel}); and the share of conditions whose outcome differs
between the two repeats, both caps pooled (Dis.). Superscripts give the stratum: B
truncation-heavy, M mild, N null. \SweepConfigs{} configurations,
\SweepRepeats{} repeats per cap, \SweepRowsKept{} kept rows; pooled rows weight
families equally.}
\label{tab:capsweep}
\resizebox{\textwidth}{!}{\setlength{\tabcolsep}{3pt}
\begin{tabular}{@{}lrrlllrr@{}}
\toprule
 & \multicolumn{2}{c}{Trunc.\ (\%)} & \multicolumn{2}{c}{Menu contrast (pp)} & & \multicolumn{1}{c}{Rel.} & \multicolumn{1}{c}{Dis.} \\
\cmidrule(lr){2-3}\cmidrule(lr){4-5}
Config.$^{\mathrm{stratum}}$ & 1{,}024 & 16{,}384 & at 1{,}024 & at 16{,}384 & Interaction & (pp) & (\%) \\
\midrule
DeepSeek V4 Pro 0813$^{B}$ & 31.9 & 0.0 & \ensuremath{-}29.5\,[\ensuremath{-}38.9,\,\ensuremath{-}20.8] & 0.7\,[\ensuremath{-}1.0,\,2.8] & 30.2\,[21.2,\,39.6] & \ensuremath{-}24.3 & 11.3 \\
GLM-5.3$^{B}$ & 69.8 & 1.4 & \ensuremath{-}61.8\,[\ensuremath{-}70.5,\,\ensuremath{-}52.8] & \ensuremath{-}2.8\,[\ensuremath{-}7.6,\,1.7] & 59.0\,[49.7,\,68.1] & \ensuremath{-}53.5 & 16.8 \\
GLM-5.3 Flash$^{B}$ & 37.5 & 0.0 & \ensuremath{-}35.1\,[\ensuremath{-}44.4,\,\ensuremath{-}25.3] & \ensuremath{-}3.8\,[\ensuremath{-}7.6,\,\ensuremath{-}0.3] & 31.2\,[21.5,\,41.0] & \ensuremath{-}28.5 & 14.6 \\
Gemini 3.8 Flash$^{B}$ & 64.9 & 0.0 & \ensuremath{-}62.8\,[\ensuremath{-}72.9,\,\ensuremath{-}52.4] & \ensuremath{-}2.1\,[\ensuremath{-}6.2,\,0.0] & 60.8\,[49.0,\,71.5] & \ensuremath{-}52.1 & 7.8 \\
Qwen3.7 Flash$^{B}$ & 51.4 & 0.0 & \ensuremath{-}46.5\,[\ensuremath{-}56.6,\,\ensuremath{-}36.5] & \ensuremath{-}4.2\,[\ensuremath{-}8.3,\,\ensuremath{-}0.7] & 42.4\,[31.9,\,53.1] & \ensuremath{-}34.7 & 14.9 \\
Qwen3.8 Flash$^{B}$ & 30.6 & 0.0 & \ensuremath{-}26.4\,[\ensuremath{-}36.5,\,\ensuremath{-}16.3] & \ensuremath{-}4.2\,[\ensuremath{-}12.2,\,2.8] & 22.2\,[13.9,\,30.6] & \ensuremath{-}16.7 & 20.7 \\
Claude Haiku 5.5$^{M}$ & 4.5 & 0.0 & \ensuremath{-}2.4\,[\ensuremath{-}6.6,\,1.7] & 0.3\,[\ensuremath{-}3.1,\,3.5] & 2.8\,[\ensuremath{-}3.1,\,8.0] & \ensuremath{-}7.6 & 7.9 \\
GPT-6 Luna$^{N}$ & 0.0 & 0.0 & \ensuremath{-}3.1\,[\ensuremath{-}8.0,\,2.4] & \ensuremath{-}2.1\,[\ensuremath{-}5.9,\,1.7] & 1.0\,[\ensuremath{-}2.4,\,4.2] & 0.0 & 8.4 \\
\midrule
Stratum B (6 cfg.) & & & \ensuremath{-}43.7\,[\ensuremath{-}48.5,\,\ensuremath{-}39.1] & \ensuremath{-}2.7\,[\ensuremath{-}4.7,\,\ensuremath{-}0.9] & 41.0\,[36.2,\,45.5] & & \\
Stratum M (1 cfg.) & & & \ensuremath{-}2.4\,[\ensuremath{-}6.6,\,1.7] & 0.3\,[\ensuremath{-}3.1,\,3.5] & 2.8\,[\ensuremath{-}3.1,\,8.0] & & \\
Stratum N (1 cfg.) & & & \ensuremath{-}3.1\,[\ensuremath{-}8.0,\,2.4] & \ensuremath{-}2.1\,[\ensuremath{-}5.9,\,1.7] & 1.0\,[\ensuremath{-}2.4,\,4.2] & & \\
All (8 cfg.) & & & \ensuremath{-}33.5\,[\ensuremath{-}37.2,\,\ensuremath{-}29.8] & \ensuremath{-}2.3\,[\ensuremath{-}3.8,\,\ensuremath{-}0.7] & 31.2\,[27.5,\,34.8] & & \\
\bottomrule
\end{tabular}
}
\end{table*}
\section{Frontier and Structural Holdout}
\label{app:frontier}

\begin{figure*}[t]
\centering
\includegraphics[width=\textwidth]{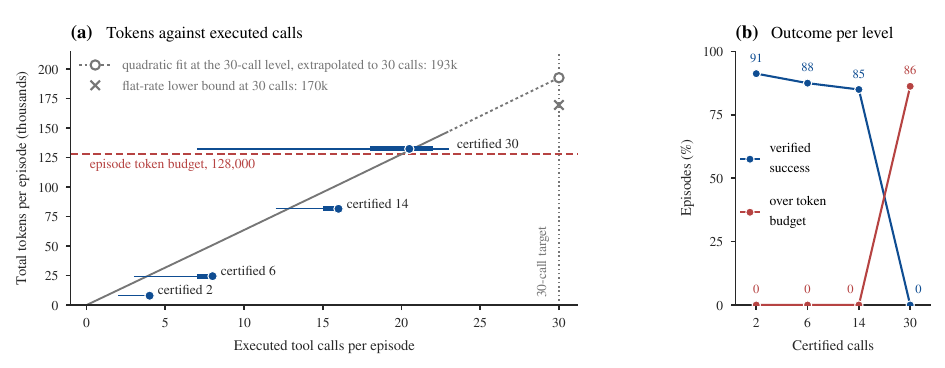}
\caption{At 30 certified calls, episodes run into the episode budget before they
run out of calls. (a) For each certified-call level, the marker is the median
number of executed calls, placed at the level's median total tokens, with the
interquartile range (thick) and the full range (thin) of executed calls. The
grey curve is the quadratic fit of tokens on attempted calls at the 30-call
level (attempted calls can exceed executed ones), solid up to the largest
attempted count and dotted beyond, ending at its
extrapolated value at 30 calls (circle) beside the flat-rate lower bound
(cross); the dashed line is the episode token budget and the dotted vertical
line the 30-call target. (b) Share of each level's episodes that were verified
successes and share that ended over the episode token budget.
\FrontierEpisodesPerLevel{} episodes per level over \FrontierModels{}
configurations and \FrontierFamilies{} families. The ranges are not
intervals.}
\label{fig:frontier-tokens}
\Description{Two panels. (a) Four summary markers for certified call levels 2,
6, 14 and 30, with horizontal range bars, against total tokens; the 30-call
marker sits just above the dashed 128,000-token budget line at about twenty
executed calls, and a fitted curve extrapolates above the budget at the dotted
30-call line. (b) Verified success stays near ninety percent at 2, 6 and 14
calls and falls to zero at 30 calls, where most episodes end over the token
budget.}
\end{figure*}

\paragraph{Frontier.}
With the \FrontierFamilies{} families as clusters, VTS changes by
\FrontierSixMinusTwo{}~pp $[\FrontierSixMinusTwoCILow, \FrontierSixMinusTwoCIHigh]$
from 2 to 6 calls and by \FrontierFourteenMinusTwo{}~pp
$[\FrontierFourteenMinusTwoCILow, \FrontierFourteenMinusTwoCIHigh]$ from 2 to 14
calls. From 2 to 30 calls it changes by \FrontierThirtyMinusTwo{}~pp
$[\FrontierThirtyMinusTwoCILow, \FrontierThirtyMinusTwoCIHigh]$, an interval
that varies only through the 2-call outcomes. At 30 calls, the
\FrontierThirtyAgentFailures{} episodes that stay within the episode budget also fail, \FrontierThirtyInfeasible{} of them at an infeasibility stop, and the
median episode executes \FrontierThirtyMedianCalls{} calls (interquartile range
\FrontierThirtyCallsIQRLow{} to \FrontierThirtyCallsIQRHigh{}) and uses
\FrontierThirtyMedianTokens{} tokens. At 14 calls the median prompt has
\FrontierFourteenMedianPromptTokens{} tokens and no episode exceeds the episode budget.
\figref{fig:frontier-tokens} shows that the 30-call episodes reach the episode
budget before they reach the call limit.

\begin{figure*}[t]
\centering
\includegraphics[width=\textwidth]{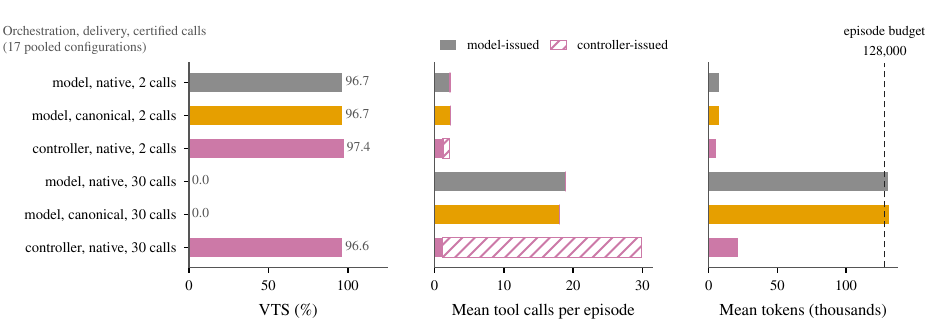}
\caption{Delegation replaces the model's calls rather than improving them. Arm
means of VTS, executed calls split by issuer, and tokens in the
structural holdout, over \StructuralModels{} configurations and
\HoldoutIntents{} held-out intents.}
\label{fig:structural}
\Description{Arm rows crossing orchestration, delivery, and certified-call level,
each with verified task success, calls split into controller-issued and
model-issued, and mean tokens.}
\end{figure*}

\paragraph{Structural holdout.}
The controller is compared with the model-run baseline, which returns records
in the cell's own encoding (native delivery), on \HoldoutIntents{}
held-out intents over \StructuralModels{} configurations (\StructuralRows{}
rows). \RegisteredStructuralModels{} configurations form the confirmatory set,
and the other \AdditionalStructuralModels{} are supplementary. Pooled over all
configurations,
$L=\StructuralAssignedLoss$~pp
$[\StructuralAssignedLossCILow, \StructuralAssignedLossCIHigh]$ and
$\Sigma=\StructuralSelectivity$~pp
$[\StructuralSelectivityCILow, \StructuralSelectivityCIHigh]$. These quantities
use the raw-encoding cell of each arm (\BaselineThirtyRows{} rows per arm),
whereas the 2-call arms of Table~\ref{tab:structural-arms} pool the raw and
compact encodings, so the arm means there differ slightly. On the confirmatory
configurations alone, $\Gamma=\RegisteredStructuralStressGain$~pp
$[\RegisteredStructuralStressGainCILow, \RegisteredStructuralStressGainCIHigh]$
and $\Delta=\RegisteredStructuralSpecificity$~pp
$[\RegisteredStructuralSpecificityCILow, \RegisteredStructuralSpecificityCIHigh]$.
The upper end of the last interval exceeds 100 because the no-repair means
cancel in $\Delta$: it is the controller's change in VTS from 2 to 30 calls minus
the control's, and so ranges over $\pm200$~pp
(Section~\ref{sec:repair-design}). The wrong-layer control, which
returns records in a canonical encoding (canonical delivery), gives
$L=\CanonicalControlAssignedLoss$~pp
$[\CanonicalControlAssignedLossCILow, \CanonicalControlAssignedLossCIHigh]$ and
$\Gamma=\CanonicalControlStressGain$~pp
$[\CanonicalControlStressGainCILow, \CanonicalControlStressGainCIHigh]$. At 2
calls the baseline reaches \BaselineTwoVTS{}\% and the controller
\ControllerTwoVTS{}\%. At 30 calls the baseline averages \BaselineThirtyCalls{}
executed calls in \BaselineThirtyRequests{} model requests. The controller
averages \ControllerThirtyCalls{} calls, \ControllerThirtyControllerCalls{} of them
issued by the controller, in \ControllerThirtyRequests{} requests
(Table~\ref{tab:structural-arms} and \figref{fig:structural}).

\begin{table*}[t]
\centering
\footnotesize
\setlength{\tabcolsep}{4pt}
\caption{Structural holdout arms. The controller succeeds at 30 calls because
it, not the model, issues the calls. Arm means over \StructuralModels{}
configurations and \HoldoutIntents{} held-out intents (\BaselineThirtyRows{}
episodes per 30-call arm): VTS, episodes over budget, executed calls by issuer,
model requests, and tokens.}
\label{tab:structural-arms}
\setlength{\tabcolsep}{3pt}
\begin{tabular}{@{}llrrrrrrrrr@{}}
\toprule
Orchestration & Delivery & Calls & Rows & VTS & Req. & Ctrl. & Model & Exec. & Tokens & Over budget \\
\midrule
Model & native & 2 & 816 & 96.7 & 3.1 & 0.0 & 2.2 & 2.2 & 7{,}227 & 0 \\
Model & canonical & 2 & 816 & 96.7 & 3.1 & 0.0 & 2.2 & 2.2 & 7{,}366 & 0 \\
Controller & native & 2 & 816 & 97.4 & 2.1 & 1.0 & 1.2 & 2.2 & 5{,}169 & 0 \\
Model & native & 30 & 408 & 0.0 & 19.8 & 0.0 & 18.9 & 18.9 & 130{,}468 & 389 \\
Model & canonical & 30 & 408 & 0.0 & 18.9 & 0.0 & 18.0 & 18.0 & 130{,}998 & 385 \\
Controller & native & 30 & 408 & 96.6 & 2.2 & 28.7 & 1.2 & 29.9 & 21{,}159 & 3 \\
\bottomrule
\end{tabular}

\end{table*}

\section{Discovery Details}
\label{app:discovery}

Discovery has \DiscoveryRows{} rows over \CoreModels{} configurations. Of these,
\DiscoveryUnavailable{} (\DiscoveryUnavailableShare{}\%) end before any model
call because the required tools were not retrieved, and
\DiscoveryAgentFailures{} are agent failures. Under flat retrieval, ITT success
falls by \RetrievalAssignedLoss{}~pp
$[\RetrievalAssignedLossCILow, \RetrievalAssignedLossCIHigh]$ from $R=32$ to
$R=2{,}048$, which is the assigned loss $L$ of the router as a repair, and the
router recovers $\Gamma=\RetrievalStressGain$~pp
$[\RetrievalStressGainCILow, \RetrievalStressGainCIHigh]$ of this loss
($\Sigma=\RetrievalSelectivity$~pp). Under
the connector-first router alone, ITT success changes by \DiscoveryRouterLoss{}~pp
$[\DiscoveryRouterLossCILow, \DiscoveryRouterLossCIHigh]$ from $R=32$ to
$R=2{,}048$.

The router is success-superior to every switch for \RetrievalSuccessSuperior{}
of \RetrievalDecisionN{} incumbents by construction: the flat lexical retriever
does not depend on the model, so every configuration gets the same availability,
and switching the downstream model cannot change it. On the
\RetrievalConditionalCases{} cases per configuration in which both retrievers
return the required tools, the router's mean success exceeds that of flat
retrieval for \RetrievalConditionalSuperior{} of the \RetrievalDecisionN{}
configurations.
\section{Failure Localization Audit}
\label{app:localization}

This audit measures how much information a failure record carries
about the assigned pressure.

\paragraph{Population and decoder.}
From the baseline cell and the three single-stress endpoint cells of the
capacity factorial, we take the \LocalizationPackets{}
configuration-family packets in which the baseline succeeds and exactly one
stressed cell fails (\LocalizationVisible{} under $V$, \LocalizationWorkflow{}
under $W$, and \LocalizationReturned{} under $D$). We exclude
\LocalizationMultiExcluded{} packets with more than one failure and
\LocalizationInfraExcluded{} with infrastructure failures. Balanced accuracy
averages the recalls of the three classes, so a constant guess scores
\BAConstant{}\%. In each of \Intents{} folds, one intent is held out, and a lookup table
maps each exact feature tuple to the majority label in the training folds. Ties
resolve in the order $V$, $W$, $D$, and unseen tuples get the training prior. No identifier of the configuration, family, or pressure enters a feature. Call counts are
binned as 0, 1 to 2, 3 to 5, and 6 or more, and trace shape adds the recorded
reason for an early stop.

\paragraph{Results.}
Table~\ref{tab:localization} gives the balanced accuracy of each feature set.
It is \BAObligation{}\% $[\BAObligationCILow, \BAObligationCIHigh]$ for the
first unmet obligation, \BATermination{}\%
$[\BATerminationCILow, \BATerminationCIHigh]$ for the terminal flag, \BAJoint{}\%
$[\BAJointCILow, \BAJointCIHigh]$ for both, \BATrace{}\%
$[\BATraceCILow, \BATraceCIHigh]$ with trace shape added, \BATraceOnly{}\%
$[\BATraceOnlyCILow, \BATraceOnlyCIHigh]$ for trace shape alone,
\BACallsOnly{}\% $[\BACallsOnlyCILow, \BACallsOnlyCIHigh]$ for call counts alone,
and \BAVerdictVector{}\% $[\BAVerdictVectorCILow, \BAVerdictVectorCIHigh]$ for
the vector of all nine obligation verdicts.
The joint record recovers \JointReturnedRecovered{} of the
\LocalizationReturned{} returned-state cases. With termination-aware labels,
balanced accuracy is \BACorrected{}\% $[\BACorrectedCILow, \BACorrectedCIHigh]$
and returned-state recall is \CorrectedReturnedRecall{}\%. A proxy that reveals
the menu size reaches \BAMenuSize{}\% $[\BAMenuSizeCILow, \BAMenuSizeCIHigh]$. It
reads the treatment, so it serves as a reference point, not a diagnostic.

\begin{table*}[t]
\centering
\footnotesize
\caption{No variant of the failure record beats call counts
alone at identifying the assigned pressure. Leave-one-intent-out balanced accuracy (\%) over
\LocalizationPackets{} single-failure packets, with 95\% intent-cluster
intervals.}
\label{tab:localization}
\begin{tabular}{@{}l l rrr r@{}}
\toprule
 & & \multicolumn{3}{c}{Recall (\%)} & Returned state \\
\cmidrule(lr){3-5}
Decoder & Balanced acc.\ [95\% CI] & $V$ & $W$ & $D$ & recovered \\
\midrule
Constant (majority) & 33.3 [33.3, 33.3] & 100.0 & 0.0 & 0.0 & 0/12 \\
Termination flag & 48.5 [43.0, 55.4] & 92.0 & 53.6 & 0.0 & 0/12 \\
First unmet obligation & 53.4 [44.6, 60.1] & 85.1 & 75.0 & 0.0 & 0/12 \\
Obligation + termination & 48.9 [42.4, 58.9] & 88.5 & 50.0 & 8.3 & 1/12 \\
+ trace shape & 51.3 [44.8, 57.0] & 93.1 & 60.7 & 0.0 & 0/12 \\
Call count only & 54.8 [47.8, 61.3] & 100.0 & 64.3 & 0.0 & 0/12 \\
Trace shape only & 52.4 [44.2, 60.2] & 96.6 & 60.7 & 0.0 & 0/12 \\
Verdict vector & 52.5 [44.3, 62.4] & 88.5 & 60.7 & 8.3 & 1/12 \\
\midrule
Corrected labels (termination-aware) & 54.1 [45.9, 63.2] & 93.1 & 60.7 & 8.3 & 1/12 \\
\midrule
Menu-size proxy (treatment-revealing) & 71.8 [60.4, 82.6] & 100.0 & 57.1 & 58.3 & 7/12 \\
\bottomrule
\end{tabular}

\end{table*}

\begin{figure*}[t]
\centering
\includegraphics[width=\textwidth]{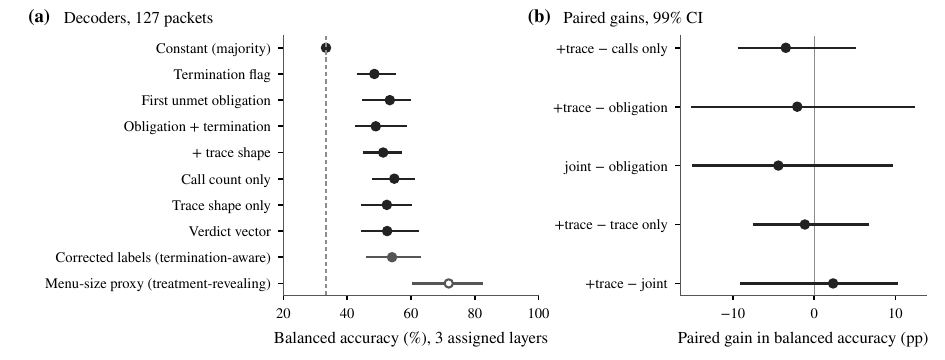}
\caption{Adding fields to the failure record does not yield a detectable gain.
(a) Balanced accuracy (\%) of each decoder on \LocalizationPackets{} packets
with 95\% intent-cluster intervals; the dashed line is the constant guess. (b)
Paired gains in balanced accuracy (pp), with 99\% Bonferroni intervals over
\Intents{} intent clusters and predictions held fixed.}
\label{fig:diagnostic-gains}
\Description{Two panels. (a) Ten decoder rows with balanced accuracy and
intervals, the constant guess marked by a dashed line. (b) Five horizontal
99 percent paired-gain intervals around a vertical zero line; all cross zero.}
\end{figure*}

We compare feature sets with paired gains and 99\% intent-cluster intervals
(Bonferroni over five exploratory comparisons, \BootstrapDraws{} draws,
out-of-fold predictions held fixed). The record with trace shape gains
\GainTraceVsCalls{}~pp $[\GainTraceVsCallsLow, \GainTraceVsCallsHigh]$ over call
counts, \GainTraceVsObligation{}~pp
$[\GainTraceVsObligationLow, \GainTraceVsObligationHigh]$ over the obligation,
\GainTraceVsTraceOnly{}~pp $[\GainTraceVsTraceOnlyLow, \GainTraceVsTraceOnlyHigh]$
over trace shape alone, and \GainTraceVsJoint{}~pp
$[\GainTraceVsJointLow, \GainTraceVsJointHigh]$ over the joint record. The joint
record gains \GainJointVsObligation{}~pp
$[\GainJointVsObligationLow, \GainJointVsObligationHigh]$ over the obligation
(\figref{fig:diagnostic-gains}). No interval excludes zero.

\paragraph{Packets with several failures.}
We also apply the decoder to \ChallengeRows{} failed stressed rows from packets
with several failures, whose intent was excluded from training. It reaches
\ChallengeBA{}\% balanced accuracy and recovers \ChallengeReturnedCorrect{} of
\ChallengeReturnedCases{} returned-state cases. The rows are correlated, so
these metrics are descriptive.
\section{Repair and Decision Details}
\label{app:repairs}

Table~\ref{tab:repairs} lists the activation quantities and the decision counts
for every repair.

\begin{table*}[t]
\centering
\footnotesize
\setlength{\tabcolsep}{3.5pt}
\caption{Routing and delegation activate, and prompt repairs do not. Only the
router (by construction) and the controller (by delegation) beat every
configuration switch on success. Assigned loss $L$, stress gain $\Gamma$, and selectivity $\Sigma$ (for the controller, wrong-layer advantage $\Delta$ in its place), in VTS pp (router: ITT success). Decision counts are
incumbents out of the configurations with complete arms. Intervals are 95\%
cluster bootstraps over \DiscoveryClusters{} required-tool clusters (router),
\SemanticFamilies{} families (routed menu and prompt repairs), or
\HoldoutIntents{} held-out intents (controller), with Holm correction over the
four layers; the controller is corrected together with its wrong-layer
control.}
\label{tab:repairs}
\resizebox{\textwidth}{!}{\setlength{\tabcolsep}{3pt}
\begin{tabular}{@{}l lll l rrr@{}}
\toprule
 & & & & (activated) & \multicolumn{3}{c}{vs.\ every switch} \\
\cmidrule(l){6-8}
Repair (layer; cfg.) & $L$ & $\Gamma$ & $\Sigma$ ($\Delta$) & Holm $p$ & Succ. & Pareto & Cost \\
\midrule
Router ($R$; 21 cfg.) & 79.6 [60.4,\,94.2] & 58.1 [36.1,\,79.9] & 58.1 [36.1,\,79.6] & .001 (yes) & 21/21 & 21/21 & 0/21 \\
Routed menu ($V$; 21 cfg.) & 17.5 [14.9,\,20.2] & 4.9 [2.6,\,7.0] & 4.9 [2.2,\,7.5] & .003 (yes) & 0/21 & 9/21 & 1/21 \\
Workflow prompt ($W$; 20 cfg.) & 2.4 [\ensuremath{-}0.4,\,5.0] & 1.7 [\ensuremath{-}0.6,\,3.8] & 0.0 [\ensuremath{-}2.4,\,2.4] & .949 (no) & 0/20 & 10/20 & 1/20 \\
Interface guide ($D$; 20 cfg.) & 1.5 [\ensuremath{-}0.1,\,3.1] & 0.1 [\ensuremath{-}1.5,\,1.9] & 0.7 [\ensuremath{-}1.7,\,3.1] & .949 (no) & 0/20 & 9/20 & 1/20 \\
\begin{tabular}[t]{@{}l@{}}Controller ($W$; 17 cfg.)\\ held-out; $\Sigma$ column: $\Delta$\end{tabular} & 96.1 [93.6,\,98.0] & 96.6 [94.6,\,98.5] & 95.3 [92.4,\,97.8] & \ensuremath{<}.001 (yes) & 17/17 & 17/17 & 4/17 \\
\bottomrule
\end{tabular}
}
\end{table*}

\paragraph{Why activation is not specificity.}
Let baseline success be $p_0>p_1$ under low and high pressure. A generic
intervention that rescues a share $r\in(0,1]$ of failures, whatever their layer,
gives success $q_z=p_z+r(1-p_z)$, so
\begin{align*}
L&=p_0-p_1>0,\\ \Gamma&=r(1-p_1)>0,\\ \Sigma&=rL>0.
\end{align*}
For $(p_0,p_1,r)=(0.8,0.4,0.5)$ this gives $L=40$, $\Gamma=30$, and $\Sigma=20$
pp without any mechanism specific to the layer. Showing specificity also
requires credible alternative repairs that are tested on the same stressed
cells and do not help.

\paragraph{Activation components.}
The routed menu has $L=\VisibleRepairAssignedLoss$~pp
$[\VisibleRepairAssignedLossCILow, \VisibleRepairAssignedLossCIHigh]$ and
$\Sigma=\VisibleRepairSelectivity$~pp
$[\VisibleRepairSelectivityCILow, \VisibleRepairSelectivityCIHigh]$. The workflow
prompt has $L=\WorkflowPromptAssignedLoss$~pp
$[\WorkflowPromptAssignedLossCILow, \WorkflowPromptAssignedLossCIHigh]$ and
$\Sigma=\WorkflowPromptSelectivity$~pp
$[\WorkflowPromptSelectivityCILow, \WorkflowPromptSelectivityCIHigh]$. The
interface guide has $L=\InterfaceGuideAssignedLoss$~pp
$[\InterfaceGuideAssignedLossCILow, \InterfaceGuideAssignedLossCIHigh]$ and
$\Sigma=\InterfaceGuideSelectivity$~pp
$[\InterfaceGuideSelectivityCILow, \InterfaceGuideSelectivityCIHigh]$. The
equivalence margins of \WorkflowPromptTOSTMargin{} and
\InterfaceGuideTOSTMargin{}~pp are the smallest symmetric margins at which
the two one-sided tests \citep{schuirmann1987comparison} reject, at the 5\% level,
a stress gain outside the margin.

\paragraph{Sentinel repeats.}
Sentinel pairs run three times and cover \SentinelGroups{} condition groups
(\SentinelRows{} rows) on \AuxModels{} configurations and \SentinelFamilies{} families. They give
\SentinelPassOne{}\% pass@1, \SentinelPassAll{}\% success on all three runs,
\SentinelPassAny{}\% success on any run, and \SentinelDisagreement{}\%
disagreement between runs. The repeated gains are \SentinelRoutedGain{}~pp
$[\SentinelRoutedGainCILow, \SentinelRoutedGainCIHigh]$ for the routed menu,
\SentinelWorkflowPromptGain{}~pp
$[\SentinelWorkflowPromptGainCILow, \SentinelWorkflowPromptGainCIHigh]$ for the
workflow prompt, and \SentinelInterfaceGuideGain{}~pp
$[\SentinelInterfaceGuideGainCILow, \SentinelInterfaceGuideGainCIHigh]$ for the
interface guide.

\paragraph{Decision matrix.}
For each incumbent, each repair is compared with every other configuration in
the same population on the same stressed cells. A repair is
success-superior if the 95\% paired bootstrap interval of its success difference
against every switch lies above zero, and cost-superior if the interval of its
cost difference against every switch lies below zero. It is Pareto-nondominated
if no switch is at least as good on success, tokens, wall time, and cost and
strictly better on one, using point estimates. The counts (success-superior,
nondominated, cost-superior) are
\RetrievalSuccessSuperior{}/\RetrievalPareto{}/\RetrievalCostSuperior{} of
\RetrievalDecisionN{} for the router,
\VisibleSuccessSuperior{}/\VisiblePareto{}/\VisibleCostSuperior{} of
\VisibleDecisionN{} for the routed menu,
\WorkflowSuccessSuperior{}/\WorkflowPareto{}/\WorkflowCostSuperior{} of
\WorkflowDecisionN{} for the workflow prompt,
\InterfaceSuccessSuperior{}/\InterfacePareto{}/\InterfaceCostSuperior{} of
\InterfaceDecisionN{} for the interface guide and
\StructuralSuccessSuperior{}/\StructuralPareto{}/\StructuralCostSuperior{} of
\StructuralModels{} for the controller.
\section{Public-Information Policy Pilot}
\label{app:pilot}

This development study asks which action to take after a failure, using only
public information. It has its own population: \PilotBlocks{} matched blocks
(\PilotEpisodes{} episodes) built from \PilotIntents{} previously inspected families, one per
intent, crossed with four cells, two incumbents, and two repeats. The cells are
$(8,\mathrm{low},\mathrm{raw})$, $(128,\mathrm{low},\mathrm{raw})$,
$(8,\mathrm{high},\mathrm{raw})$, and $(8,\mathrm{low},\mathrm{compact})$. The
incumbents are Qwen3.8 Flash and DeepSeek V4 Flash 0731 on the crossover's
route, and the fixed switch target is GPT-5.6 Luna, a high-baseline
configuration (Table~\ref{tab:panel}). 

\paragraph{Actions and access.}
Each block contains a probe and seven restarts, each reset independently: an
unchanged retry, a \CrossoverCapHigh{}-token completion cap, a public-description
router, the public-handoff controller, canonical delivery (the wrong-layer encoding control), the combined repairs
(router, controller, and canonical delivery with the larger cap), and the fixed switch. The router matches a unique
route in the user text to public tool descriptions before the first request. It
never receives the gold operational tools, and it leaves the menu unchanged when
the route is absent or ambiguous. A trace selector chooses an action from public
numeric features of requests and responses only. It is a depth-one tree fitted
with each intent held out.

\paragraph{Results.}
The unchanged retry succeeds on \PilotRetryBlocks{} of \PilotBlocks{} blocks
(\PilotRetry{}\%), the public router on \PilotRouterBlocks{} (\PilotRouter{}\%),
the combined repairs on \PilotBundleBlocks{} (\PilotBundle{}\%), the trace selector on
\PilotTraceBlocks{} (\PilotTrace{}\%), and the fixed switch on
\PilotSwitchBlocks{} (\PilotSwitch{}\%) (Table~\ref{tab:policy-pilot}). The
difference between the trace selector and the switch is
\PilotTraceDifference{}~pp $[\PilotTraceDifferenceLow, \PilotTraceDifferenceHigh]$
over intent clusters, with the fitted rules held fixed. With the probe charged,
the mean cost per 1,000 conditions at frozen prices is
USD~\PilotTraceCostPerThousand{} for trace selection and
USD~\PilotSwitchCostPerThousand{} for probing and then switching, against
USD~\PilotSwitchNoProbeCostPerThousand{} for switching without a probe.

\paragraph{Conditional on a failed probe.}
\PilotSuccessfulProbes{} of the \PilotBlocks{} probes succeed, so the pooled
rates above mix recovery with restarts that were not needed. Among the
\PilotFailedProbes{} blocks whose probe failed, the unchanged retry recovers
\PilotRetryRecovered{} (\PilotRetryRecoveredPct{}\%), the larger cap
\PilotBudgetRecoveredPct{}\%, the router \PilotRouterRecovered{}
(\PilotRouterRecoveredPct{}\%), the combined repairs \PilotBundleRecovered{}
(\PilotBundleRecoveredPct{}\%), and the switch \PilotSwitchRecovered{}
(\PilotSwitchRecoveredPct{}\%). The policy a practitioner would run, keeping a
successful probe and acting only after a failure, reaches
\PilotBundlePolicyPct{}\% with the combined repairs and \PilotSwitchPolicyPct{}\% with the
switch (Table~\ref{tab:policy-pilot}, after-failed-probe columns).

\begin{table*}[t]
\centering
\footnotesize
\caption{In the development pilot, a fixed switch beats every public repair.
Block-level success (\%) and cost per action at frozen prices, over
\PilotBlocks{} blocks and two incumbents, on \PilotIntents{} previously inspected families,
one per intent. LOIO: leave-one-intent-out.}
\label{tab:policy-pilot}
\resizebox{\textwidth}{!}{\setlength{\tabcolsep}{3pt}
\begin{tabular}{@{}l r rr rr rr@{}}
\toprule
 & \multicolumn{3}{c}{Success, \% (blocks)} & \multicolumn{2}{c}{After failed probe} & \multicolumn{2}{c}{USD per 1{,}000 conditions} \\
\cmidrule(lr){2-4}\cmidrule(lr){5-6}\cmidrule(l){7-8}
Action or policy & Pooled & DeepSeek V4 Flash 0731 & Qwen3.8 Flash & Recovered & Retain-then-act & With probe & No probe \\
\midrule
Retain incumbent (probe) & 66.1 (127/192) & 49.0 (47/96) & 83.3 (80/96) & 0.0 (0/65) & 66.1 (127/192) & 2.68 &  \\
Unchanged retry & 67.7 (130/192) & 52.1 (50/96) & 83.3 (80/96) & 24.6 (16/65) & 74.5 (143/192) & 5.38 & 2.70 \\
Larger completion cap & 73.4 (141/192) & 55.2 (53/96) & 91.7 (88/96) & 41.5 (27/65) & 80.2 (154/192) & 5.76 & 3.09 \\
Public-description router & 78.6 (151/192) & 66.7 (64/96) & 90.6 (87/96) & 61.5 (40/65) & 87.0 (167/192) & 4.14 & 1.46 \\
Public-handoff controller & 71.9 (138/192) & 56.2 (54/96) & 87.5 (84/96) & 35.4 (23/65) & 78.1 (150/192) & 5.24 & 2.56 \\
Canonical delivery & 71.4 (137/192) & 54.2 (52/96) & 88.5 (85/96) & 35.4 (23/65) & 78.1 (150/192) & 5.56 & 2.88 \\
Combined repairs & 81.8 (157/192) & 67.7 (65/96) & 95.8 (92/96) & 66.2 (43/65) & 88.5 (170/192) & 4.24 & 1.56 \\
Fixed switch (GPT-5.6 Luna) & 99.0 (190/192) & 99.0 (95/96) & 99.0 (95/96) & 98.5 (64/65) & 99.5 (191/192) & 4.62 & 1.94 \\
Trace selector (LOIO) & 97.9 (188/192) & 97.9 (94/96) & 97.9 (94/96) &  &  & 4.49 &  \\
\bottomrule
\end{tabular}
}
\end{table*}
\section{External Environment}
\label{app:external}

\begin{figure*}[t]
\centering
\includegraphics[width=\textwidth]{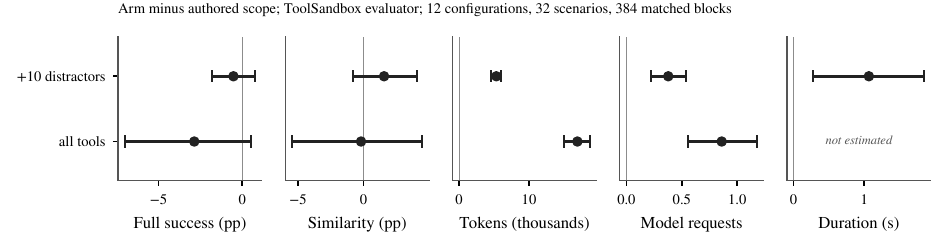}
\caption{In ToolSandbox, a larger menu costs tokens more clearly than success.
Pooled paired contrasts of each menu arm (ten added distractors; all tools)
against the authored scope on ToolSandbox's own evaluator: full success and
similarity (pp), tokens, model requests, and duration, over
\ConcurrentMenuModels{} concurrent configurations, \MenuScenarios{} scenarios, and
\ConcurrentMenuBlocks{} matched blocks, with 95\% scenario-cluster intervals. All-tools duration is not
estimated. The historical stratum is confounded with time and reported
separately in Table~\ref{tab:toolsandbox}.}
\label{fig:toolsandbox}
\Description{Five dot-and-interval panels, one per metric, each with a row for
the distractor arm and a row for the all-tools arm; the success and similarity
intervals cross zero, while the token and request intervals lie above zero.}
\end{figure*}

We test whether the menu effect carries over to ToolSandbox
\citep{lu2025toolsandbox}, a stateful benchmark scored by its own evaluator
rather than by VTS. In the concurrent menu
stratum (\ConcurrentMenuModels{} configurations, \MenuScenarios{} scenarios,
\ConcurrentMenuBlocks{} matched blocks, \ConcurrentMenuRows{} rows), exposing all
tools instead of the authored scope changes full success by
\MenuAllToolsSuccess{}~pp $[\MenuAllToolsSuccessCILow, \MenuAllToolsSuccessCIHigh]$
(sign-flip $p=\MenuAllToolsSuccessP$) and similarity by
\MenuAllToolsSimilarity{}~pp
$[\MenuAllToolsSimilarityCILow, \MenuAllToolsSimilarityCIHigh]$, while adding
\MenuAllToolsTokens{} tokens $[\MenuAllToolsTokensCILow, \MenuAllToolsTokensCIHigh]$
and \MenuAllToolsRequests{} requests
$[\MenuAllToolsRequestsCILow, \MenuAllToolsRequestsCIHigh]$. This arm has no
duration estimate because its episodes are continuation-only segments
(Table~\ref{tab:toolsandbox}). Ten added distractors change
full success by \MenuTenSuccess{}~pp
$[\MenuTenSuccessCILow, \MenuTenSuccessCIHigh]$, similarity by
\MenuTenSimilarity{}~pp $[\MenuTenSimilarityCILow, \MenuTenSimilarityCIHigh]$,
tokens by \MenuTenTokens{} $[\MenuTenTokensCILow, \MenuTenTokensCIHigh]$, and duration by \MenuTenDuration{}~s
$[\MenuTenDurationCILow, \MenuTenDurationCIHigh]$
(Table~\ref{tab:toolsandbox} and \figref{fig:toolsandbox}). In ToolSandbox, a
larger menu thus costs tokens more clearly than success.

A separate historical stratum (\HistoricalMenuModels{} configurations), run at a
different time, gives a similarity contrast of
\HistoricalMenuStressContrast{}~pp
$[\HistoricalMenuStressContrastCILow, \HistoricalMenuStressContrastCIHigh]$ under
stress. The two strata are never pooled. Across \ExternalRankModels{}
configurations, the rank correlation between ToolSandbox success and VTS is
$\rho=\ExternalRho$, with intervals of
$[\ExternalRhoCILow, \ExternalRhoCIHigh]$ when resampling scenarios and families and
$[\ExternalRhoConfigCILow, \ExternalRhoConfigCIHigh]$ when resampling
configurations (Kendall $\tau_b=\ExternalKendall$). Both intervals lie below
zero, so over these \ExternalRankModels{} configurations VTS and ToolSandbox
success are negatively rank-correlated.

\begin{table*}[t]
\centering
\footnotesize
\caption{In ToolSandbox, a larger menu costs tokens more clearly than success.
Paired scenario-level differences under ToolSandbox's evaluator (success and
similarity in pp, tokens, requests, and seconds) for \ConcurrentMenuModels{}
concurrent configurations and \MenuScenarios{} scenarios, with 95\%
scenario-cluster intervals.}
\label{tab:toolsandbox}
\setlength{\tabcolsep}{3pt}
\begin{tabular}{@{}p{0.29\linewidth} p{0.32\linewidth} p{0.32\linewidth}@{}}
\toprule
Concurrent menu (12 cfg., 32 scenarios, 384 blocks) & +10 distractors $-$ scope & All tools $-$ scope \\
\midrule
Full success (pp) & \ensuremath{-}0.5 [\ensuremath{-}1.8, 0.8] & \ensuremath{-}2.9 [\ensuremath{-}7.0, 0.5] \\
Similarity (pp) & 1.6 [\ensuremath{-}0.8, 4.0] & \ensuremath{-}0.2 [\ensuremath{-}5.5, 4.5] \\
Tokens per episode & 5{,}322 [4{,}622, 6{,}018] & 16{,}916 [15{,}021, 18{,}676] \\
Model requests & 0.38 [0.22, 0.54] & 0.86 [0.56, 1.18] \\
Duration (s) & 1.07 [0.28, 1.85] & \emph{not estimated}: continuation-only segment is not a complete episode duration \\
\midrule
Historical menu (5 cfg.): stress-arm similarity contrast (pp) & \multicolumn{2}{l}{\ensuremath{-}0.7 [\ensuremath{-}5.2, 3.9]} \\
External rank (5 cfg.): Spearman $\rho$ & \multicolumn{2}{l}{\ensuremath{-}0.8 [\ensuremath{-}1.0, \ensuremath{-}0.5] scenario/family; [\ensuremath{-}1.0, \ensuremath{-}0.11] configuration} \\
External rank: Kendall $\tau_b$ & \multicolumn{2}{l}{\ensuremath{-}0.6} \\
\bottomrule
\end{tabular}

\end{table*}

\section{Comparison with Related Benchmarks}
\label{app:neighbors}

\begin{table*}[t]
\centering
\small
\setlength{\tabcolsep}{4pt}
\caption{Ten design properties across related benchmarks.
\ding{51}~=~property documented in the abstract or full text,
\textsf{p}~=~partial or adjacent, \textendash~=~not found.
Columns: FBG~=~FuncBenchGen \citep{maekawa2026funcbenchgen},
TMB~=~ToolMenuBench \citep{babu2026toolmenubench},
AFT~=~AFT-Bench \citep{wang2026aftbench},
TRB~=~ToolRobustBench \citep{zheng2026toolrobustbench},
AFB~=~APIFlow-Bench \citep{wan2026apiflowbench},
HB~=~Harness-Bench \citep{yao2026harnessbench},
MoH~=~\emph{Model or Harness?} \citep{raj2026modelorharness},
LRB~=~LayerRAG-Bench \citep{shams2026layerragbench},
CAF~=~CAFE \citep{lukassen2026cafe},
TSB~=~ToolSandbox \citep{lu2025toolsandbox},
TBB~=~ToolBottleneckBench (this work).}
\label{tab:neighbors}
\begin{tabular}{@{}lccccccccccc@{}}
\toprule
Property & FBG & TMB & AFT & TRB & AFB & HB & MoH & LRB & CAF & TSB & TBB \\
\midrule
Same-goal controlled contrast   & \ding{51} & \ding{51} & \ding{51} & \ding{51} & \ding{51} & \ding{51} & \textendash & \ding{51} & \ding{51} & \ding{51} & \ding{51} \\
Separate availability estimand  & \textsf{p} & \ding{51} & \textsf{p} & \textendash & \textendash & \textendash & \textendash & \textsf{p} & \textendash & \textendash & \ding{51} \\
Crossed factors within one goal & \ding{51} & \ding{51} & \textsf{p} & \textsf{p} & \textendash & \ding{51} & \textendash & \ding{51} & \ding{51} & \textendash & \ding{51} \\
Certified stateful horizon      & \textsf{p} & \textsf{p} & \textsf{p} & \textendash & \ding{51} & \textsf{p} & \textendash & \textsf{p} & \textendash & \textsf{p} & \ding{51} \\
Observation-format factor       & \textsf{p} & \textendash & \ding{51} & \ding{51} & \textsf{p} & \textendash & \textendash & \ding{51} & \textendash & \textendash & \ding{51} \\
Exact executable grading        & \ding{51} & \textsf{p} & \ding{51} & \ding{51} & \ding{51} & \textsf{p} & \textendash & \textsf{p} & \textendash & \textsf{p} & \ding{51} \\
Matched-layer repair test       & \ding{51} & \ding{51} & \ding{51} & \ding{51} & \textendash & \textendash & \textendash & \ding{51} & \textendash & \textendash & \ding{51} \\
Wrong-layer control             & \textendash & \textendash & \textsf{p} & \ding{51} & \textendash & \textendash & \textendash & \ding{51} & \textendash & \textendash & \ding{51} \\
Model-vs-system decision        & \textendash & \textendash & \textendash & \textendash & \textendash & \ding{51} & \ding{51} & \textendash & \ding{51} & \textendash & \ding{51} \\
Repeated-run stochasticity      & \textendash & \textendash & \textsf{p} & \textendash & \ding{51} & \textendash & \textendash & \textendash & \textsf{p} & \textsf{p} & \textsf{p} \\
\bottomrule
\end{tabular}

\vspace{2pt}
{\footnotesize\raggedright
\textbf{Sources for the non-obvious cells.}
FBG separate availability is \textsf{p} because its distractor functions are
always visible, with no registered-versus-visible split. Its matched repair is
restating prior variable values, which the authors report lifts GPT-5 from
62.5\% to 81.3\%.
TMB fills the availability row by measuring gold next-tool exposure before
selection separately from downstream wrong-tool calls and goal-state success.
AFT holds the task, backend, initial state, injected failure, agent, and model
fixed while varying the interface. Its wrong-layer cell is \textsf{p} because
each mechanism is evaluated only under its own matched treatment.
TRB reports two distinct intervention sets: alias repair on A-family tasks moves
success from 21/30 to 30/30, while chain-of-thought prompting on B-family tasks
leaves 21/30 unchanged despite diagnosed selection failures.
AFB uses an oracle admission screen and reports five epochs per task, with
all-five reliability spanning 44 points against 7 for best-of-five.
HB combines deterministic validators where available with rubric-based judgment
elsewhere (its Section~3.4), which supports partial grading and certification
marks.
MoH develops a failure-localization taxonomy with a judge-agreement study rather
than controlled task-repair effects. Its repair-assignment framing supports the
model-vs-system row.
LRB fills both repair rows: schema normalization lifts schema-drift success from
0.000 to 0.913 and does not succeed on the off-target layers it lists.
CAF is scored on a shared rubric with a configurable LLM judge, so its
exact-grading cell is \textendash. It fills the model-vs-system row through
mixed-effects attribution of variance to components.
TSB exposes 0, 3, 10, or all remaining tools as distractors, prioritizing domain
overlap and textual similarity, and grades by milestones and minefields. Its
four-trial user-simulator error audit supports a partial repeated-run mark.
TBB's repeated-run cell is \textsf{p}: three-repeat sentinel pairs cover
\AuxModels{} configurations in \SentinelFamilies{} families, and the two-cap
crossover and the cap sweep repeat the factorial twice on \CrossoverConfigs{}
and \SweepConfigs{} configurations. The
core factorial has one realization per cell.\par}
\end{table*}

Table~\ref{tab:neighbors} compares ToolBottleneckBench (TBB) with related benchmarks.
Every property appears in at least one prior design, but no prior benchmark has more than five of the ten;
ToolBottleneckBench has nine and repeated-run stochasticity in part.
TwinCheck's counterfactual \emph{negative twins} \citep{dai2026twincheck} replace a tool call at inference
time; our matched variants are compiled before the episode.
\section{Reproducibility, Verification, and Cost}
\label{app:repro}

\paragraph{Availability.}
\ifdefined\BottleneckPreprint
The public code and data repository is \url{https://github.com/zichuanxu/tool-bottleneck-bench}.
\else
An anonymized copy of the release repository is provided at \url{https://anonymous.4open.science/r/TBB-0307}.
\fi
It contains the deterministic generator, the exact grader, every scored row, the
frozen plans, the analysis code, and the manifests. Running new hosted episodes
requires provider credentials. The sweep and policy-pilot analyses rerun offline
from the released rows; the whole-panel check and the certificates are released
with their passed reports. Authored code and data are released under the MIT
License; ToolSandbox keeps its upstream license, and model-generated records
remain subject to the providers' terms.

\paragraph{Verification.}
For \CertificateModels{} configurations, certificates reconstruct the frozen
harness verdicts from saved responses. ToolSandbox rows are re-graded from saved
state, without running the tools again. A whole-panel check re-runs the same
analysis code deterministically, reproduces every typed field of the
\RecomputedAnalyses{} recomputed analyses, and carries \PreservedAnalyses{}
analyses byte for byte. The scoped test suite passes
\TestsPassed{} tests and \SubtestsPassed{} subtests (\JUnitCases{} cases).

\paragraph{Cost.}
Applying frozen route prices to normalized token counts, without cache credit,
gives a cost of USD~\TotalCost{} for the \TotalRows{} rows. Table~\ref{tab:panel} gives the cost of each configuration, and
the release gives its token counts. The most expensive configuration is
\CostMostExpensiveModel{} (USD~\CostMostExpensiveUSD{}) and the cheapest
\CostCheapestModel{} (USD~\CostCheapestUSD{}).

\end{document}